\documentclass[12pt]{spieman}  % 12pt font required by SPIE;
\usepackage{amsmath,amsfonts,amssymb}
\usepackage{graphicx}
\usepackage{setspace}
\usepackage{tocloft}
\usepackage{epstopdf}
\usepackage{algorithm}
\usepackage{algorithmic}
\usepackage{enumerate}
\usepackage{enumitem}
\usepackage{bm}
\usepackage{amsmath}
\usepackage{url}
\usepackage{booktabs}
\usepackage{multirow}

\title{Total Variation with Overlapping Group Sparsity and Lp Quasinorm  for Infrared Image Deblurring under Salt-and-Pepper Noise}

\author[a,b]{Xingguo Liu}
\author[a,c]{Yinping Chen}
\author[a,*]{Zhenming Peng}
\author[b]{Juan Wu}
\affil[a]{University of Electronic Science and Technology of China, School of Information and Communication Engineering, Chengdu, China, 610054}
\affil[b]{Chongqing College of Electronic Engineering, Chongqing, China, 401331}
\affil[c]{Minnan Normal University, School of Physics and Information Engineering, Zhangzhou, China, 363000}

\cftpagenumbersoff{figure}
\cftpagenumbersoff{table}
\begin{document}
\maketitle

\begin{abstract}
Because of the limitations of the infrared imaging principle and the properties of infrared imaging systems, infrared images have some drawbacks, including a lack of details, indistinct edges, and a large amount of salt-and-pepper noise. To improve the sparse characteristics of the image while maintaining the image edges and weakening staircase artifacts, this paper proposes a method that uses the Lp quasinorm instead of the L1 norm and for infrared image deblurring with an overlapping group sparse total variation method. The Lp quasinorm introduces another degree of freedom, better describes image sparsity characteristics, and improves image restoration. Furthermore, we adopt the accelerated alternating direction method of multipliers and fast Fourier transform theory in the proposed method to improve the efficiency and robustness of our algorithm. Experiments show that under different conditions for blur and salt-and-pepper noise, the proposed method leads to excellent performance in terms of objective evaluation and subjective visual results.
\end{abstract}

% Include a list of up to six keywords after the abstract
\keywords{infrared image; overlapping group sparsity; Lp quasinorm; salt-and-pepper noise deblurring; accelerated alternating multiplier iterative method}

% Include email contact information for corresponding author
{\noindent \footnotesize\textbf{*}Zhenming Peng,  \linkable{zmpeng@uestc.edu.cn} }

\begin{spacing}{2}   % use double spacing for rest of manuscript

\section{Introduction}
\label{sect:intro}  % \label{} allows reference to this section
Infrared images have the characteristics of high levels of background noise and low resolution. The target of an infrared imaging system is often situated against a complex background and there is a low signal-to-noise ratio. The target occupies a small number of pixels on the imaging surface, and the low resolution results in a lack of sufficient information, such as details and shape features, which makes the detection of the target difficult. Therefore, the enhancement of infrared images and techniques to suppress noise are key tasks and ongoing challenges in the field of infrared image processing research.

As the main source of noise in infrared imaging system, the detector has a complicated mechanism and is the main factor affecting the image quality of infrared systems. The noise of the detector itself is unavoidable. According to the mechanism that produces it, noise can be divided into thermal noise, shot noise, photon noise, and other types. The part of the noise that has a large influence on the image can be considered to be equivalent to Gaussian white noise and salt-and-pepper noise. In addition, during image capture, degradation of the observed image can be caused by various factors such as defocusing, diffraction, relative motion between the detector and the object.

Image restoration is the improvement of the quality of a degraded image. It removes or mitigates the degradation of the image quality that occurs during the acquisition of the digital image in order to visually improve the image. The most typical degradation phenomena are blur and noise. This paper mainly discusses the restoration of blurry images, that is, deblurring.

The blurring process of the image can be expressed as a convolution of the original image with the blur kernel and superimposed noise, that is $g = h*f + n$, where $*$ is the convolution operator, $g$ denotes a blurred image containing noise, $f$ denotes the original image, $h$ is a blur kernel, also called a point spread function (PSF), and $n$ is noise. The inverse processing of a blurred image is called image deconvolution, and its purpose is to recover a clear image from the blurred image. According to whether the PSF is known, the image deconvolution problem is divided into two types: blind deconvolution and non-blind deconvolution.

Non-blind image deconvolution assumes that both the blurred image and blur kernel for estimating a clear image have been given. In image restoration processing, non-blind image deconvolution is an ill-conditioned inverse problem that is often modeled by the regularization method as the following energy function minimization model:
\begin{equation}\label{eq1}
\begin{aligned}
\underset{f}{\mathop{\min }}\,\frac{1}{2}\left\| h*f-g \right\|_{2}^{2}+\mu \psi \left( f \right)\
\end{aligned},
\end{equation}
Here, the first term is the data fidelity term and the second term is the regularization term (alternatively the constraint term or regularization function); the regularization parameter is used to control the weighted ratio between the fidelity term and the regularization term. Different regularization methods are generated depending on the regularization terms. The earliest regularization method is the Tikhonov regularization method, proposed by Tikhonov et al. in 1977 \cite{RN627} and its regularization term in the image deblurring problem is $\psi (f)=\left\| \nabla f \right\|_{2}^{2}$. The regularization term can effectively suppress noise, but often produces a smooth image, so that the processing result is still blurred. To overcome the shortcomings of the Tikhonov regularization method, Rudin et al. proposed the total variation (TV) regularization method\cite{RN360}, and its regularization term in the image deblurring problem is $\psi (f)={{\left\| \nabla f \right\|}_{1}}$. The TV regularization method can suppress noise and preserve the edges of the image, but it can only effectively approximate the slice constant function, so staircase artifacts are often generated in the smooth image regions, reducing the image restoration quality. To reduce the staircase artifacts of the restored image and preserve its edge information, Lysaker et al. proposed a second-order TV regularization to replace the original TV regularization term \cite{RN576}. Chan et al. proposed a hybrid TV method that combines first-order and second-order TV \cite{RN628}. Luo et al. proposed a weighted difference of anisotropic TV (ATV) and isotropic TV (ITV) model for image processing\cite{lou2015weighted}. Other researchers have also proposed image restoration models based on high-order TV regularization terms. Although they can effectively suppress staircase artifacts, the detailed information and important features of the image are often unclear \cite{RN629,RN630}. Huang et al. proposed the fast TV (Fast-TV) minimization method by introducing auxiliary variables \cite{RN625}. Bredies et al. proposed total generalized variation (TGV) to replace the commonly used TV regularization term. The TGV image restoration model effectively approximates a polynomial function of any order, and can effectively suppress noise during image restoration while protecting the important details of the image and improving the quality of image restoration \cite{RN284}.

However, when the image is affected by impulse noise, the TGV image restoration model cannot recover important information from the degraded image because the assumptions are no longer correct. To address the statistical characteristics of impulse noise, the TV model based on the L1 data fidelity term was proposed to restore images with this type of noise \cite{RN368,RN281}. It is expressed as follows.
\begin{equation}\label{eq2}
\begin{aligned}
\underset{f}{\mathop{\min }}\,\frac{1}{2}{{\left\| h*f-g \right\|}_{1}}+\mu \psi \left( f \right)\
\end{aligned},
\end{equation}

Similarly, because the TV regularization term can only effectively approximate a piecewise constant function, the resulting staircase artifacts tend to reduce the image restoration quality. Hence, many researchers have proposed improved versions of TV. The non-local total variation (NLTV) model\cite{RN642,RN643} can suppress staircase artifacts and preserve the detailed information of the image, but it is too computationally complex to use in to practical engineering problems. The studies \cite{RN644,RN385} proposed the L1-high-order TV (L1-HTV) model, which can effectively reduce the staircase artifacts in smooth image regions, but cannot effectively protect the important details of the image. Liu et al. used overlapping groups of sparse regularizations to recover noise-damaged images \cite{RN407}. This method is very effective at reducing staircase artifacts. Bai et al. proposed a model based on the overlap direction multiplier method to solve TV regularization. This model is very effective for removing salt-and-pepper, but it is not effective for removing random noise \cite{RN609}.

In recent years, Selesnick and Chen proposed overlapping group sparse TV (OGSTV) \cite{RN407,RN485,RN623}. The regularization term is a non-separating regularization term that better preserves the sparsity of the objective function \cite{RN624}. The overlapping group sparse regularization term not only considers the sparsity of the image difference domain, but also obtains the neighborhood difference information of each point, thus determining the structural sparsity characteristics of the image gradient. By overlapping the combined gradients, the difference between smooth regions and boundary regions can be improved, thereby suppressing the staircase artifacts of the TV model. Based on the work of Selesnick and Chen, Liu et al. extended the one-dimensional OGS regularization term into a two-dimensional OGS regularization term and introduced it into an ATV model for the denoising and deconvolution of images with salt-and-pepper noise based on the L1 norm (OGSATVL1) \cite{RN365}. In addition, Liu et al. used OGS regularization terms for speckle noise removal \cite{RN402}.

In the traditional model, the TV is based on the L1 norm, but in practice, many non-convex reconstruction models are better than the L1 norm-based sparse constrained reconstruction model at low sampling rates. Yuan and Ghanem proposed a new sparse optimization method for impulse noise image restoration called L0TVPADMM, which solves the TV-based restoration problem with L0-norm data fidelity and solves the method using a proximal Alternating Direction Method of Multipliers (PADMM)\cite{yuan2015l0tv}. Chartrand et al. first proposed a non-convex optimization problem using Lp norm minimization ($0<p<1$) as the objective function \cite{RN299,RN401}. Later, Chartrand and Staneva collaborated to give theoretical Lp-reconfigurable conditions for arbitrary sparse signals \cite{RN632}. Wu et al. \cite{RN633} and Wen et al. \cite{RN634} further theoretically demonstrated the superiority of the Lp norm-based method. Compared with the L1 norm, since the Lp norm is non-convex and non-smooth in the case of $0<p<1$, the solution is more complicated. At present, there are three main algorithms for solving Lp norm-based problems: the iterative weighted L1 algorithm \cite{RN635}, iterative reweighting least squares method \cite{RN636}, and iterative threshold algorithm \cite{RN637}. In particular, for the iterative threshold algorithm based on Lp norm-based problems, it has been confirmed that when $p = 1/2 $ or  $2/3$, the expression for the threshold can be explicitly given \cite{RN354,RN638}. Xu and colleagues have deeply and meticulously researched many theoretical and applied aspects of the case in which $p = 1/2$ \cite{RN637,RN638,RN640}.

In this study, we explore the Lp quasinorm relaxation to improve the sparsity exploitation of OGSTV; our proposed method is referred to as the OGSTV with Lp quasinorm (OGSTVLp), which is efficiently solved through alternating direction method of multipliers (ADMM) in conjunction with non-convex p-shrinkage mapping. The novelty of our work is three-fold. First, the OGSTVLp method is far less restrictive than the OGSTV method for infrared image reconstruction; it not only shows good performance in terms of detail preservation, but also achieves accurate measurement of the sparsity potential from the regularity prior. Second, an efficient iterative algorithm is proposed to optimize the ADMM with a fast and stable convergence result. Third, fast and efficient closed-form solutions are investigated and derived for computationally complex sub-minimization problems using fast Fourier transforms (FFT).

The remainder of this paper is organized as follows. Section 2 briefly introduces the OGSTV method, the majorization minimization (MM) method, and Sparse TV Based on the Lp quasinorm method. Section 3 describes the proposed method as well as the fast ADMM algorithm. Then, in Section 4, our experiments and results are described. Finally, Sections 5 and 6 present our discussion and conclusions, respectively.

\section{Preliminaries}
\label{sect:Preliminaries}  % \label{} allows reference to this section
\subsection{OGSTV}

When the additive noise in an image is salt-and-pepper noise, because of its sparsity, the data fidelity term adopts the L1 norm. The resulting OGSTV deblurring model is as follows:
\begin{equation}\label{eq3}
\begin{aligned}
\bm F =\arg \underset{\bm F}{\mathop{\min }}\,\frac{1}{2}{{\left\| \bm H*\bm F-\bm G \right\|}_{1}}+\mu {{\text{R}}_{OGSTV}}\left( \bm F \right)\
\end{aligned},
\end{equation}
where $*$ indicates the convolution operator, $ \bm F\in {{R}^{N\times N}} $ indicates the restored image, $ \bm G\in {{R}^{N\times N}}$ indicates the blurred and noisy image, $ \frac{1}{2}{{\left\| \bm H*\bm F-\bm G \right\|}_{1}}$ represents the fidelity term, ${{\text{R}}_{OGSTV}}\left( \bm F \right)$ represents the OGSTV regularization term, $\mu $ is the coefficient balancing the fidelity and the OGSTV regularization, and ${{\left\| \centerdot  \right\|}_{1}}$ reprents the L1 norm, defined as ${{\left\| \bm F \right\|}_{1}}\text{=}\sum\limits_{i=1}^{N}{\sum\limits_{j=1}^{N}{\left| F_{ij}^{{}} \right|}}$.

Term ${{\text{R}}_{OGSTV}}\left( \bm F \right)$ is defined as follows:
\begin{equation}\label{eq4}
\begin{aligned}
{{\text{R}}_{\text{OGSTV}}}\left( \bm F \right)\text{=}\varphi ({{\bm K}_{h}}*\bm F)\text{+}\varphi ({{\bm K}_{v}}*\bm F)
\end{aligned},
\end{equation}
where ${{\bm K}_{h}}\text{=}[-1,1]$, ${{\bm K}_{v}}\text{=}\left[ \begin{matrix}
   -1  \\
   1  \\
\end{matrix} \right]$ represents the horizontal and vertical differential convolution kernels, respectively. Moreover, $\varphi (\bm V)$ indicates the overlapping group gradient of the processed pixel, which is defined as follows:
\begin{equation}\label{eq5}
\begin{aligned}
\varphi (\bm V)=\sum\limits_{\text{i}=1}^{N}{\sum\limits_{\text{j}=1}^{N}{{{\left\| {{\overset{\sim}{\mathop{V}}\,}_{i,j,K,K}} \right\|}_{2}}}}
\end{aligned},
\end{equation}
where ${{\overset{\sim}{\mathop{V}}\,}_{\text{i,j,K,K}}}$  is  the overlapping group matrix, which is further defined as follows:
\begin{equation}\label{eq6}
\begin{aligned}
{{\overset{\sim}{\mathop{V}}\,}_{i,j,K,K}}\text{=}\left[ \begin{matrix}
   {{V}_{i-{{K}_{l}},j-{{K}_{l}}}} & {{V}_{i-{{K}_{l}},j-{{K}_{l}}+1}} & \cdots  & {{V}_{i-{{K}_{l}},j\text{+}{{K}_{r}}}}  \\
   {{V}_{i-{{K}_{l}}\text{+}1,j-{{K}_{l}}}} & {{V}_{i-{{K}_{l}}\text{+}1,j-{{K}_{l}}+1}} & \cdots  & {{V}_{i-{{K}_{l}}\text{+}1,j\text{+}{{K}_{r}}}}  \\
   \vdots  & \vdots  & \ddots  & \vdots   \\
   {{V}_{i+{{K}_{r}},j-{{K}_{l}}}} & {{V}_{i+{{K}_{r}},j-{{K}_{l}}+1}} & \cdots  & {{V}_{i+{{K}_{r}},j+{{K}_{r}}}}  \\
\end{matrix} \right]\in {{R}^{K\times K}}
\end{aligned},
\end{equation}
where ${{K}_{l}}=\left\lfloor \frac{K-1}{2} \right\rfloor $, ${{K}_{r}}=\left\lfloor \frac{K}{2} \right\rfloor $, and $\left\lfloor \text{x} \right\rfloor $ denotes the largest integer less than or equal to $\text{x}$.

Equation \ref{eq6} shows that OGSTV considers the neighborhood gradient information for an overlapping group matrix of $ K \times K $ points. In this way, the similarity of the neighborhood structure is fully explored to improve the difference between the high-noise points of the smooth region and the pixel of the boundary region, thereby denoising the image more robustly.

\subsection{the majorization minimization method}
The optimization algorithm can be used to solve the overlap group sparse denoising model by minimization as follows:
\begin{equation}\label{eq7}
\begin{aligned}
P(\bm V)=pro{{x}_{\gamma \varphi }}({{\bm V}_{0}})=\underset{\bm V}{\mathop{\arg \min }}\,\frac{1}{2}\left\| \bm V-{{\bm V}_{0}} \right\|_{2}^{2}+\gamma \varphi (\bm V).
\end{aligned}
\end{equation}

According to the majorization minimization (MM )method, the iterated solution is as follows:
\begin{equation}\label{eq8}
\begin{aligned}
{{\bm V}^{(k+1)}}=\underset{\bm V}{\mathop{\arg \min }}\,Q(\bm V,{{\bm V}^{(k)}}),
\end{aligned}
\end{equation}
where $\textbf I\in {{\mathbb{R}}^{{{N}^{2}}\times {{N}^{2}}}}$ is the unit matrix, ${{\bm v}_{0}}$ is the vector form of ${{V}_{0}}$ , and  reshapes a vector into a matrix. Matrix $\bm D(\bm U)\in {{\mathbb{R}}^{{{N}^{2}}\times {{N}^{2}}}}$ is a diagonal matrix in which each diagonal component is
\begin{equation}\label{eq9}
\begin{aligned}
{{\left[ \bm D(\bm U) \right]}_{m,m}}=\sqrt{\sum\limits_{i=-{{K}_{l}}}^{{{K}_{r}}}{\sum\limits_{j=-{{K}_{l}}}^{{{K}_{r}}}{{{\left\{ \sum\limits_{{{k}_{1}}=-{{K}_{l}}}^{{{K}_{r}}}{\sum\limits_{{{k}_{2}}=-{{K}_{l}}}^{{{K}_{r}}}{{{\left| {{U}_{m-i+{{k}_{1}},m-j+{{k}_{2}}}} \right|}^{2}}}} \right\}}^{-\frac{1}{2}}}}}}.
\end{aligned}
\end{equation}

Therefore, we obtain Algorithm 1 to solve the Eq.\ref{eq7}.

\begin{algorithm}[htbp!]
\caption{MM method}
\label{algorithm1}
\vspace*{0mm}
\textbf{Initialize:} $v={{v}_{0}}$, $\gamma $, ${{K}^{2}}$, ${{K}_{l}}=\left[ \frac{K-1}{2} \right]$, ${{K}_{r}}=\left[ \frac{K}{2} \right]$, $\varepsilon $, \text {Maximum inner iterations} $NIt$,$k=0$  \\
\textbf{While} ${{{\left\| {{\bm V}^{(k+1)}}-{{\bm V}^{(k)}} \right\|}_{2}}}/{{{\left\| {{\bm V}^{(k)}} \right\|}_{2}}}\;>\varepsilon$ \text{or} $k<NIt$     \textbf{do}\\
\vspace*{-5mm}
\begin{enumerate}\label{enumi1}
\item ${{\left[ \bm D_{{}}^{2}({{\bm V}^{(k)}}) \right]}_{m,m}}=\sum\limits_{i=-{{K}_{l}}}^{{{K}_{r}}}{\sum\limits_{j=-{{K}_{l}}}^{{{K}_{r}}}{{{\left\{ \sum\limits_{{{k}_{1}}=-{{K}_{l}}}^{{{K}_{r}}}{\sum\limits_{{{k}_{2}}=-{{K}_{l}}}^{{{K}_{r}}}{{{\left| \bm V_{m-i+{{k}_{1}},m-j+{{k}_{2}}}^{(k)} \right|}^{2}}}} \right\}}^{-\frac{1}{2}}}}}$
\item ${{\bm V}^{(k+1)}}=mat\left\{ {{\left(\textbf I+\gamma {{\bm D}^{2}}\left( {{\bm V}^{(k)}} \right) \right)}^{-1}}{{v}_{0}} \right\}$
\item $k=k+1$
\end{enumerate}
\textbf{End While} \\
\textbf{Return ${{\bm V}^{(k)}}$}
\end{algorithm}

\subsection{Sparse TV Based on the Lp Quasinorm}
\begin{figure}
\begin{center}
\begin{tabular}{c}
\includegraphics[width=11cm]{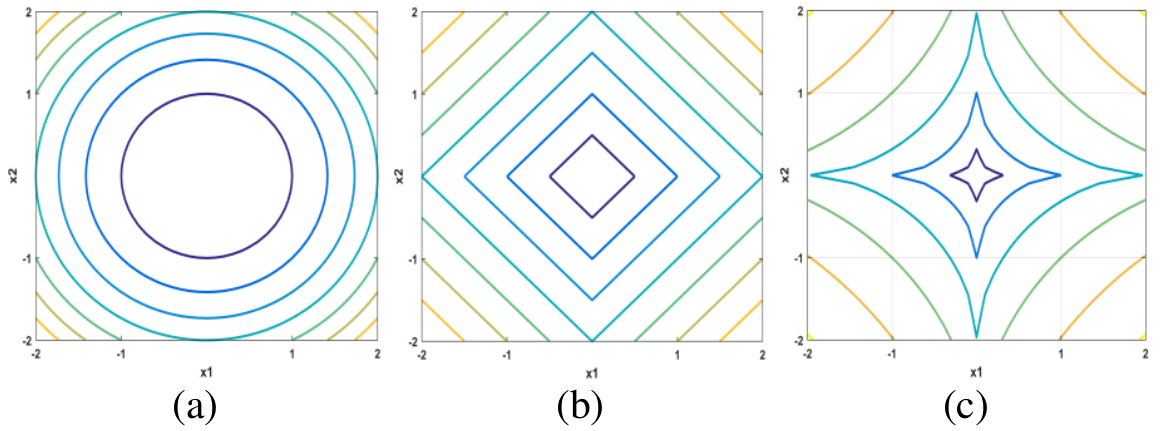}
\\
%(a) \hspace{2.9cm} (b) \hspace{2.9cm} (c)
\end{tabular}
\end{center}
\caption
{ \label{fig1}
The contour line of Lp-quasinorm: (a) $p=2$, (b) $p=1$ and (c) $0<p<1$.}
\end{figure}

Compared with the L1 and L2 norms, the Lp quasinorm has one more degree of freedom; therefore, it can better characterize sparse gradient information. The contours of the ATV ${{R}_{ApTV}}(\bm F)=\left\| {{\bm K}_{h}}*\bm F \right\|_{p}^{p}+\left\| {{\bm K}_{v}}*\bm F \right\|_{p}^{p}\begin{array}{*{35}{l}}
   (0<p\le 2) \end{array} $  based on the Lp quasinorm are shown in Fig.\ref{fig1}, where the L1 and L2 norms are special cases of the Lp norm. The Lp norm is defined as ${{\left\| \bm F \right\|}_{p}}=(\sum\limits_{i=1}^{N}{\sum\limits_{j=1}^{N}{{{\left| {{\bm F}_{ij}} \right|}^{p}}{{)}^{{1}/{p}\;}}}}$, while the Lp quasinorm is defined as $\left\| \bm F \right\|_{p}^{p}=\sum\limits_{i=1}^{N}{\sum\limits_{j=1}^{N}{{{\left| {{\bm F}_{ij}} \right|}^{p}}}}$. 
%   In Fig.\ref{fig1}, white Gaussian noise is added to the image with a standard deviation of $\sigma $.  As can be seen from the figure, the intersection of the L2 norm contours and fidelity term $\frac{1}{2}\left\| \bm F-\bm G \right\|_{2}^{2}$ is not sparse, whereas the intersection of the L1 norm contours and fidelity term is sparse, but is susceptible to noise pollution. In contrast, the contours of the Lp quasinorm are more robust to noise.
As can be seen from the figure, the smaller the parameter p is, the sparser the solution domain of Lp-quasinorm is.

%Hence, the ATV based on the L1 norm is extended to $ \text T{{\text V}_{p}}(\bm F)$  as follows:
%\begin{equation}\label{eq10}
%\text {TV}_p(\bm F)=\| \nabla_1 \bm F\|_p^p+\| \nabla_2 \bm F\|_p^p, \quad (0<p<1),
%\end{equation}
%where $ \text {TV}_{p}(\bm F)$ represents the ATV sparse regularization term based on the Lp quasinorm.

\section{Proposed method}
\label{sect:Proposed method}  % \label{} allows reference to this section
In this study, we propose a deblurring method for infrared images based on OGSTV with the Lp quasinorm, which we call OGSTVLp. It is expressed as follows:
\begin{equation}\label{eq11}
\bm F=\underset{\bm F}{\mathop{\arg \min }}\,\varphi ({{\bm K}_{h}}*\bm F)+\varphi ({{\bm K}_{v}}*\bm F)+\mu \left\| \bm H*\bm F-\bm G \right\|_{p}^{p},0<p<1.
\end{equation}

To solve the OGSTVLp model in the framework of ADMM, some additional variables are required to convert the unconstrained problem given by Eq.\ref{eq11} into the following constrained problem:
\begin{equation}\label{eq12}
\begin{split}
(\bm F,\bm T,\bm W,{{\bm Z}_{1}},{{\bm Z}_{2}})=&\underset{\bm F,\bm T,\bm W,{{\bm Z}_{1}},{{\bm Z}_{2}}}{\mathop{\arg \min }}\,\varphi ({{\bm Z}_{1}})\text{+}\varphi ({{\bm Z}_{2}})+\mu \left\| \bm W \right\|_{p}^{p} \\
 =&\underset{\bm F,\bm T,\bm W,{{\bm Z}_{1}},{{\bm Z}_{2}}}{\mathop{\arg \min }}\,\sum\limits_{i=1}^{2}{\varphi ({{\bm Z}_{i}})}+\mu \left\| \bm W \right\|_{p}^{p}, \\
 & s.t.   {{\bm Z}_{1}}\text{=}{{\bm K}_{h}}*\bm F,  {{\bm Z}_{2}}\text{=}{{\bm K}_{v}}*\bm F,  \bm W=\bm H*\bm F-\bm G,   \bm T=\bm F.
\end{split}
\end{equation}

Consequently, the corresponding augmented Lagrangian function is as follows:
\begin{equation}\label{eq13}
\begin{split}
&\mathcal{L}({{\bm Z}_{1}},{{\bm Z}_{2}},\bm W,\bm T,\bm F;{{\bm V}_{1}},{{\bm V}_{2}},{{\bm V}_{3}},{{\bm V}_{4}}) \\
 & =\sum\limits_{i=1}^{2}{\varphi ({{\bm Z}_{i}})}-\sum\limits_{i=1}^{2}{\left\langle {{\bm V}_{i}},({{\bm Z}_{i}}-{{\bm K}_{i}}*\bm F) \right\rangle }+\frac{{{\lambda }_{1}}}{2}\sum\limits_{i=1}^{2}{\left\| {\bm {Z}_{i}}-{{\bm K}_{i}}*F \right\|_{2}^{2}} \\
 & +\mu \left\| \bm W \right\|_{p}^{p}-\left\langle {{\bm V}_{3}},(\bm W-(\bm H*\bm F-\bm G)) \right\rangle +\frac{{{\lambda }_{2}}}{2}\left\| \bm W-(\bm H*\bm F-\bm G) \right\|_{2}^{2} \\
 & -\left\langle {{\bm V}_{4}},\bm T-\bm F \right\rangle +\frac{{{\lambda }_{3}}}{2}\left\| \bm T-\bm F \right\|_{2}^{2},
\end{split}
\end{equation}
where ${{\bm V}_{i}}(i=1,2,3,4)$ is a Lagrange multiplier and ${{\lambda }_{i}}>0,(i=1,2,3)$ is a penalty parameter.

The minimizer of Eq. \ref{eq12} is the saddle point of $\mathcal{L}({{\bm Z}_{1}},{{\bm Z}_{2}},\bm W,\bm T,\bm F;{{\bm V}_{1}},{{\bm V}_{2}},{{\bm V}_{3}},{{\bm V}_{4}})$, which can be found by solving the following sequence of subproblems:
\begin{equation}\label{eq14}
\begin{split}
   \bm Z_{i}^{\left( k+1 \right)}=&\underset{{{\bm Z}_{i}}}{\mathop{\arg \min }}\,\varphi ({{\bm Z}_{i}})-\left\langle \bm V_{i}^{\left( k \right)},({{\bm Z}_{i}}-{{\bm K}_{i}}*{{\bm F}^{\left( k \right)}}) \right\rangle +\frac{{{\lambda }_{1}}}{2}\left\| {{\bm Z}_{i}}-{{\bm K}_{i}}*{{\bm F}^{\left( k \right)}} \right\|_{2}^{2} \\
  =&\underset{{{\bm Z}_{i}}}{\mathop{\arg \min }}\,\varphi ({{\bm Z}_{i}})+\frac{{{\lambda }_{1}}}{2}\left\| {{\bm Z}_{i}}-{{\bm K}_{i}}*{{\bm F}^{\left( k \right)}}-\frac{\bm V_{i}^{\left( k \right)}}{{{\lambda }_{1}}} \right\|_{2}^{2},\begin{matrix}
   {} & i=1,2  \\
\end{matrix}
\end{split}
\end{equation}
\begin{equation}\label{eq15}
\begin{split}
  {{\bm W}^{\left( k+1 \right)}}=&\underset{\bm W}{\mathop{\arg \min }}\,\mu \left\| \bm W \right\|_{p}^{p}-\left\langle \bm V_{3}^{\left( k \right)},(\bm W-(\bm H*{{\bm F}^{\left( k \right)}}-\bm G)) \right\rangle \\
  +& \frac{{{\lambda }_{2}}}{2}\left\| \bm W-(\bm H*{{\bm F}^{\left( k \right)}}-\bm G) \right\|_{2}^{2} \\
  =& \underset{\bm W}{\mathop{\arg \min }}\,\mu \left\| \bm W \right\|_{p}^{p}+\frac{{{\lambda }_{2}}}{2}\left\| \bm W-(\bm H*{{\bm F}^{\left( k \right)}}-\bm G)-\frac{\bm V_{3}^{\left( k \right)}}{{{\lambda }_{2}}} \right\|_{2}^{2}
\end{split}
\end{equation}
\begin{equation}\label{eq16}
\begin{split}
  {{\bm T}^{\left( k+1 \right)}}=&\underset{\bm T}{\mathop{\arg \min }}\,-\left\langle \bm V_{4}^{\left( k \right)},\bm T-{{\bm F}^{\left( k \right)}} \right\rangle +\frac{{{\lambda }_{3}}}{2}\left\| \bm T-{{\bm F}^{\left( k \right)}} \right\|_{2}^{2} \\
 =&\underset{\bm T}{\mathop{\arg \min }}\,\frac{{{\lambda }_{3}}}{2}\left\| \bm T-{{\bm F}^{\left( k \right)}}-\frac{V_{4}^{\left( k \right)}}{{{\lambda }_{3}}} \right\|_{2}^{2}
\end{split}
\end{equation}
\begin{equation}\label{eq17}
\begin{split}
  {{\bm F}^{\left( k+1 \right)}}=&\sum\limits_{i=1}^{2}{\left\langle \bm V_{i}^{\left( k+1 \right)},(\bm Z_{i}^{\left( k+1 \right)}-{{\bm K}_{i}}*\bm F) \right\rangle }+\frac{{{\lambda }_{1}}}{2}\sum\limits_{i=1}^{2}{\left\| \bm Z_{i}^{\left( k+1 \right)}-{{\bm K}_{i}}*\bm F \right\|_{2}^{2}} \\
 -& \left\langle \bm V_{3}^{\left( k+1 \right)},({{\bm W}^{\left( k+1 \right)}}-(\bm H*\bm F-\bm G)) \right\rangle +\frac{{{\lambda }_{2}}}{2}\left\| {{\bm W}^{\left( k+1 \right)}}-(\bm H*\bm F-\bm G) \right\|_{2}^{2} \\
 -& \left\langle \bm V_{4}^{\left( k+1 \right)},{{\bm T}^{\left( k+1 \right)}}-\bm F \right\rangle +\frac{{{\lambda }_{3}}}{2}\left\| {{\bm T}^{\left( k+1 \right)}}-\bm F \right\|_{2}^{2} \\
 =& \frac{{{\lambda }_{1}}}{2}\sum\limits_{i=1}^{2}{\left\| \bm Z_{i}^{\left( k+1 \right)}-{{\bm K}_{i}}*\bm F-\frac{\bm V_{i}^{\left( k+1 \right)}}{{{\lambda }_{1}}} \right\|_{2}^{2}} \\
 +& \frac{{{\lambda }_{2}}}{2}\left\| {{\bm W}^{\left( k+1 \right)}}-(\bm H*\bm F-\bm G)-\frac{\bm V_{3}^{\left( k+1 \right)}}{{{\lambda }_{2}}} \right\|_{2}^{2} \\
 +& \frac{{{\lambda }_{3}}}{2}\left\| {{\bm T}^{\left( k+1 \right)}}-\bm F-\frac{\bm V_{4}^{\left( k+1 \right)}}{{{\lambda }_{3}}} \right\|_{2}^{2},\begin{matrix}
   {} & i=1,2.  \\
\end{matrix}
\end{split}
\end{equation}

The procedure consists of the following steps:

1. To solve the sub-problem of ${{\bm Z}_{i}}$ in Eq.\ref{eq14}, the MM (Algorithm \ref{algorithm1}) can be used.

2. The sub-problem $\bm W$ in Eq.\ref{eq15} can be solved using a soft threshold operator as follows:
\begin{equation}\label{eq18}
{{\bm W}^{(k+1)}}=shrin{{k}_{p}}\left( \left( \bm H*{{\bm F}^{(k)}}-\bm G \right)+\frac{\bm V_{3}^{(k)}}{{{\lambda }_{2}}},\frac{\mu }{{{\lambda }_{2}}} \right),
\end{equation}
where
\begin{equation}\label{eq19}
shrin{{k}_{p}}\left( \xi ,\frac{1}{\beta } \right)=\max \left\{ \left| \xi  \right|-{{\beta }^{p-2}}{{\left| \xi  \right|}^{p-1}},0 \right\}\cdot \frac{\xi }{\left| \xi  \right|}.
\end{equation}

3. In Eq.\ref{eq16}, the minimizer is given explicitly by
\begin{equation}\label{eq20}
{{\bm T}^{(k+1)}}={{P}_{\Omega }}\left( {{\bm F}^{(k)}}+\frac{\bm V_{4}^{(k)}}{{{\lambda }_{3}}} \right),
\end{equation}
where ${{P}_{\Omega }}$ is defined as the projection operator on the set $\Omega =\left\{ \bm F\in {{\mathbb{R}}^{N\times N}}|0\le \bm F\le 1 \right\}$ as
\begin{equation}\label{eq21}
{{P}_{\Omega }}{{\left( F \right)}_{i,j}}=\left\{ \begin{array}{*{35}{l}}
   0, & {{F}_{i,j}}<0;  \\
   {{F}_{i,j}}, & {{F}_{i,j}}\in [0,1];  \\
   1, & {{F}_{i,j}}>1;  \\
\end{array} \right.
\end{equation}

4. By employing the convolution theorem, the two-dimensional Fourier transform of $\bm F$ in Eq.\ref{eq17} can be obtained as follows:
\begin{equation}\label{eq22}
\begin{split}
&\mathcal{\bm F}\left( {{\bm F}^{(k+1)}} \right)={{\lambda }_{1}}\sum\limits_{i=1}^{2}{{{\left[ \mathcal F\left( {{\bm K}_{i}} \right) \right]}^{*}}\circ \left( \mathcal F\left( {{\bm K}_{i}} \right)\circ \mathcal F\left( \bm F \right)+\frac{\mathcal F\left( \bm V_{i}^{\left( k+1 \right)} \right)}{{{\lambda }_{1}}}-\mathcal F\left( \bm Z_{i}^{\left( k+1 \right)} \right) \right)} \\
 & +{{\lambda }_{2}}{{\left[ \mathcal F\left( \bm H \right) \right]}^{*}}\circ \left( (\mathcal F\left( \bm H \right)\circ \mathcal F\left( \bm F \right)-\mathcal F\left( \bm G \right))+\frac{\mathcal F\left( \bm V_{3}^{\left( k+1 \right)} \right)}{{{\lambda }_{2}}}-\mathcal F\left( {{\bm W}^{\left( k+1 \right)}} \right) \right) \\
 & +{{\lambda }_{3}}\left( \mathcal F\left( \bm F \right)+\frac{\mathcal F\left( \bm V_{4}^{\left( k+1 \right)} \right)}{{{\lambda }_{3}}}-\mathcal F\left( {{\bm T}^{\left( k+1 \right)}} \right) \right),
 \end{split}
\end{equation}
When we set $\mathcal F\left( {{\bm F}^{(k+1)}} \right)=0 $, it can be resolved as follows:
\begin{equation}\label{eq23}
\bm {lhs}={{\lambda }_{1}}\sum\limits_{i=1}^{2}{{{\left[ \mathcal F\left( {{\bm K}_{i}} \right) \right]}^{*}}\circ \mathcal F\left( {{\bm K}_{i}} \right)+{{\lambda }_{2}}{{\left[ \mathcal F\left(\bm H \right) \right]}^{*}}\circ \mathcal F\left(\bm H \right)+{{\lambda }_{3}}\textbf I},
\end{equation}
\begin{equation}\label{eq24}
\begin{split}
\bm {rhs} &={{\lambda }_{1}}\sum\limits_{i=1}^{2}{{{\left[ \mathcal F\left( {{\bm K}_{i}} \right) \right]}^{*}}\circ \left(\mathcal F\left( \bm Z_{i}^{\left( k+1 \right)} \right)-\frac{\mathcal F\left( \bm V_{i}^{\left( k+1 \right)} \right)}{{{\lambda }_{1}}} \right)} \\
 & +{{\lambda }_{2}}{{\left[\mathcal F\left( \bm H \right) \right]}^{*}}\circ \left(\mathcal F\left( \bm G \right)+\mathcal F\left( {{\bm W}^{\left( k+1 \right)}} \right)-\frac{\mathcal F\left( \bm V_{3}^{\left( k+1 \right)} \right)}{{{\lambda }_{2}}} \right) \\
 & +{{\lambda }_{3}}\left(\mathcal F\left( {{\bm T}^{\left( k+1 \right)}} \right)-\frac{\mathcal F\left( \bm V_{4}^{\left( k+1 \right)} \right)}{{{\lambda }_{3}}} \right)
\end{split}
\end{equation}
\begin{equation}\label{eq25}
{{\bm F}^{(k+1)}}={{\mathcal F}^{-1}}\left( \bm {rhs}\cdot /\bm {lhs} \right).
\end{equation}

5. We then update the multiplier as
\begin{equation}\label{eq26}
\begin{split}
\left\{
\begin{array}{*{35}{l}}
   \bm V_{1}^{\left( k+1 \right)}=\bm V_{1}^{\left( k \right)}-\gamma {{\lambda }_{1}}\left( \bm Z_{1}^{\left( k+1 \right)}-{{\bm K}_{1}}*{{\bm F}^{\left( k+1 \right)}} \right)  \\
   \bm V_{2}^{\left( k+1 \right)}=\bm V_{2}^{\left( k \right)}-\gamma {{\lambda }_{1}}\left( \bm Z_{2}^{\left( k+1 \right)}-{{\bm K}_{2}}*{{\bm F}^{\left( k+1 \right)}} \right)  \\
   \bm V_{3}^{\left( k+1 \right)}=\bm V_{3}^{\left( k \right)}-\gamma {{\lambda }_{2}}\left( \bm W_{{}}^{\left( k+1 \right)}-\left( \bm H*{{\bm F}^{\left( k+1 \right)}}-\bm G \right) \right)  \\
   \bm V_{4}^{\left( k+1 \right)}=\bm V_{4}^{\left( k \right)}-\gamma {{\lambda }_{3}}\left( \bm T_{{}}^{\left( k+1 \right)}-{{\bm F}^{\left( k+1 \right)}} \right)  \\
\end{array} \right.
\end{split}
\end{equation}

The proposed method is summarized in Algorithm \ref{algorithm2}.
\begin{algorithm}[htbp!]
\caption{OGSATVLp-ADMM}
\label{algorithm2}
\vspace*{0mm}
\textbf{Initialize:} $\bm Z_{1}^{(0)}=\bm Z_{2}^{(0)}=\bm G$, $k=0$, ${{\lambda }_{1}}$, ${{\lambda }_{2}}$, ${{\lambda }_{3}}$, $\gamma$, $\mu $, group size ${{K}^{2}}$, $\bm V_{i}^{(0)}=0,i=1,2,3,4$, maximum inner iterations  $NIt$ \\
\textbf{Iterate}   \\
\vspace*{-5mm}
\begin{enumerate}\label{enumi2}
\item compute $\bm Z_{1}^{(k+1)}$ and $\bm Z_{2}^{(k+1)}$  according to Eq.\ref{eq14}
\item compute ${{\bm W}^{(k+1)}}$ according to Eq.\ref{eq18}
\item compute ${{\bm T}^{(k+1)}}$ according to Eq.\ref{eq20}
\item compute ${{\bm F}^{(k+1)}}$ according to Eqs.\ref{eq23},\ref{eq24},\ref{eq25}
\item update  $\bm V_{i}^{\left( k+1 \right)},i=1,2,3,4$ according to Eq.\ref{eq26}
\item $k=k+1$
\end{enumerate}
\textbf{Until a stopping criterion is satisfied.} \\
\end{algorithm}

In addition, from the fast ADMM algorithm proposed by Goldstein et al.\cite{RN388}, we adopt the accelerated step $\alpha _{i}^{(k+1)}$, variables $\tilde{\bm Z}_{1}^{(k+1)},\tilde{\bm Z}_{2}^{(k+1)}$ and dual variables $\tilde{\bm V}_{1}^{(k+1)},\tilde{\bm V}_{2}^{(k+1)}$, expressed respectively as follows:
\begin{equation}\label{eq27}
\alpha _{i}^{(k+1)}=\frac{1+\sqrt{1+4{{(\alpha _{i}^{(k)})}^{2}}}}{2},
\end{equation}
\begin{equation}\label{eq28}
\tilde{\bm Z}_{1}^{(k+1)}=\bm Z_{1}^{(k+1)}+\frac{\alpha _{1}^{(k)}-1}{\alpha _{1}^{(k+1)}}\left( \bm Z_{1}^{(k+1)}-\bm Z_{1}^{(k)} \right),
\end{equation}
\begin{equation}\label{eq29}
\tilde{\bm Z}_{2}^{(k+1)}=\bm Z_{2}^{(k+1)}+\frac{\alpha _{2}^{(k)}-1}{\alpha _{2}^{(k+1)}}\left( \bm Z_{2}^{(k+1)}-\bm Z_{2}^{(k)} \right),
\end{equation}
\begin{equation}\label{eq30}
\tilde{\bm W}_{{}}^{(k+1)}=\bm W_{{}}^{(k+1)}+\frac{\alpha _{3}^{(k)}-1}{\alpha _{3}^{(k+1)}}\left(\bm W_{{}}^{(k+1)}-\bm W_{{}}^{(k)} \right),
\end{equation}
\begin{equation}\label{eq31}
\tilde{\bm T}_{{}}^{(k+1)}=\bm T_{{}}^{(k+1)}+\frac{\alpha _{4}^{(k)}-1}{\alpha _{4}^{(k+1)}}\left( \bm T_{{}}^{(k+1)}-\bm T_{{}}^{(k)} \right),
\end{equation}
\begin{equation}\label{eq32}
\tilde{\bm V}_{i}^{(k+1)}=\bm V_{i}^{(k+1)}+\frac{\alpha _{i}^{(k)}-1}{\alpha _{i}^{(k+1)}}\left( \bm V_{i}^{(k+1)}-\bm V_{i}^{(k)} \right).
\end{equation}

Then, the Eqs.\ref{eq14},\ref{eq18},\ref{eq20},\ref{eq24},\ref{eq26} are expressed respectively as follows:
\begin{equation}\label{eq33}
\bm Z_{i}^{\left( k+1 \right)}=\underset{{{\bm Z}_{i}}}{\mathop{\arg \min }}\,\varphi ({{\bm Z}_{i}})+\frac{{{\lambda }_{1}}}{2}\left\| {{\bm Z}_{i}}-{{\bm K}_{i}}*{{\bm F}^{\left( k \right)}}-\frac{\tilde{\bm V}_{i}^{\left( k \right)}}{{{\lambda }_{1}}} \right\|_{2}^{2}
\end{equation}
\begin{equation}\label{eq34}
{{\bm W}^{(k+1)}}=shrin{{k}_{p}}\left( \left(\bm H*{{\bm F}^{(k)}}-\bm G \right)+\frac{\tilde{\bm V}_{3}^{(k)}}{{{\lambda }_{2}}},\frac{\mu }{{{\lambda }_{2}}} \right)
\end{equation}
\begin{equation}\label{eq35}
{{\bm T}^{(k+1)}}={{P}_{\Omega }}\left( {{\bm F}^{(k)}}+\frac{\tilde{\bm V}_{4}^{(k)}}{{{\lambda }_{3}}} \right)
\end{equation}
\begin{equation}\label{eq36}
\begin{split}
\bm {rhs} &= {{\lambda }_{1}}\sum\limits_{i=1}^{2}{{{\left[\mathcal F\left( {{\bm K}_{i}} \right) \right]}^{*}}\circ \left(\mathcal F\left( \bm Z_{i}^{\left( k+1 \right)} \right)-\frac{F\left( \tilde{\bm V}_{i}^{\left( k+1 \right)} \right)}{{{\lambda }_{1}}} \right)} \\
    &+{{\lambda }_{2}}{{\left[ \mathcal F\left( \bm H \right) \right]}^{*}}\circ \left( \mathcal F\left( \bm G \right)+\mathcal F\left( {{\bm W}^{\left( k+1 \right)}} \right)-\frac{\mathcal F\left( \tilde{\bm V}_{3}^{\left( k+1 \right)} \right)}{{{\lambda }_{2}}} \right) \\
    &+{{\lambda }_{3}}\left( \mathcal F\left( {{\bm T}^{\left( k+1 \right)}} \right)-\frac{\mathcal F\left( \tilde{\bm V}_{4}^{\left( k+1 \right)} \right)}{{{\lambda }_{3}}} \right)
\end{split}
\end{equation}
\begin{equation}\label{eq37}
\begin{split}
\left\{
\begin{array}{*{35}{l}}
   \bm V_{1}^{\left( k+1 \right)}=\tilde{\bm V}_{1}^{\left( k \right)}-\gamma {{\lambda }_{1}}\left( \bm Z_{1}^{\left( k+1 \right)}-{{\bm K}_{1}}*{{\bm F}^{\left( k+1 \right)}} \right)  \\
   \bm V_{2}^{\left( k+1 \right)}=\tilde{\bm V}_{2}^{\left( k \right)}-\gamma {{\lambda }_{1}}\left( \bm Z_{2}^{\left( k+1 \right)}-{{\bm K}_{2}}*{{\bm F}^{\left( k+1 \right)}} \right)  \\
   \bm V_{3}^{\left( k+1 \right)}=\tilde{\bm V}_{3}^{\left( k \right)}-\gamma {{\lambda }_{2}}\left( \bm W_{{}}^{\left( k+1 \right)}-\left( \bm H*{{\bm F}^{\left( k+1 \right)}}-\bm G \right) \right)  \\
   \bm V_{4}^{\left( k+1 \right)}=\tilde{\bm V}_{4}^{\left( k \right)}-\gamma {{\lambda }_{3}}\left( \bm T_{{}}^{\left( k+1 \right)}-{{\bm F}^{\left( k+1 \right)}} \right)  \\
\end{array} \right.
\end{split}
\end{equation}

This algorithm is summarized in Algorithm \ref{algorithm3}.
\begin{algorithm}[htbp!]
\caption{OGSATVLp-Fast ADMM}
\label{algorithm3}
\vspace*{0mm}
\textbf{Initialize:} $\bm Z_{1}^{(0)}=\bm Z_{2}^{(0)}=\bm G$, $k=0$, ${{\lambda }_{1}}$, ${{\lambda }_{2}}$, ${{\lambda }_{3}}$, $\gamma$, $\mu $, group size ${{K}^{2}}$, $\bm V_{i}^{(0)}=0,i=1,2,3,4$, maximum inner iterations  $NIt$ \\
\textbf{Iterate}   \\
\vspace*{-5mm}
\begin{enumerate}\label{enumi3}
\item compute $\bm Z_{1}^{(k+1)}$ and $\bm Z_{2}^{(k+1)}$  according to Eq.\ref{eq33}
\item compute ${{\bm W}^{(k+1)}}$ according to Eq.\ref{eq34}
\item compute ${{\bm T}^{(k+1)}}$ according to Eq.\ref{eq35}
\item compute ${{\bm F}^{(k+1)}}$ according to Eqs.\ref{eq23},\ref{eq36},\ref{eq25}
\item update  $\bm V_{i}^{\left( k+1 \right)},i=1,2,3,4$ according to Eq.\ref{eq37}
\item compute $d_{i}^{(k+1)}={{\left( \gamma {{\lambda }_{1}} \right)}^{-1}}{{\left\|\bm V_{i}^{(k+1)}-\tilde{\bm V}_{i}^{(k+1)} \right\|}^{2}}+\gamma {{\lambda }_{1}}{{\left\|\bm Z_{i}^{(k+1)}-\tilde{\bm Z}_{i}^{(k+1)} \right\|}^{2}}$
\item \textbf {if} $d_{i}^{(k+1)}<\eta d_{i}^{(k)}$, \textbf {then}
\item compute $\alpha _{i}^{(k+1)}$ according to Eq.\ref{eq27}
\item compute $\tilde{\bm Z}_{i}^{(k+1)}$ according to Eq.\ref{eq28}
\item compute $\tilde{\bm V}_{i}^{(k+1)}$ according to Eq.\ref{eq32}
\item \textbf {else}
\item $\alpha _{i}^{(k+1)}=1$, $\tilde{\bm Z}_{i}^{(k+1)}=\bm Z_{i}^{(k+1)}$, $\tilde{\bm V}_{i}^{(k+1)}=\bm V_{i}^{(k+1)}$, $d_{i}^{(k+1)}={{\eta }^{-1}}d_{i}^{(k)}$
\item \textbf {endif}
\item $k=k+1$
\end{enumerate}
\textbf{Until a stopping criterion is satisfied.} \\
\end{algorithm}

\section{Experiments and results}
\subsection{Data and parameters}
\begin{figure}[htbp!]
\begin{center}
\includegraphics{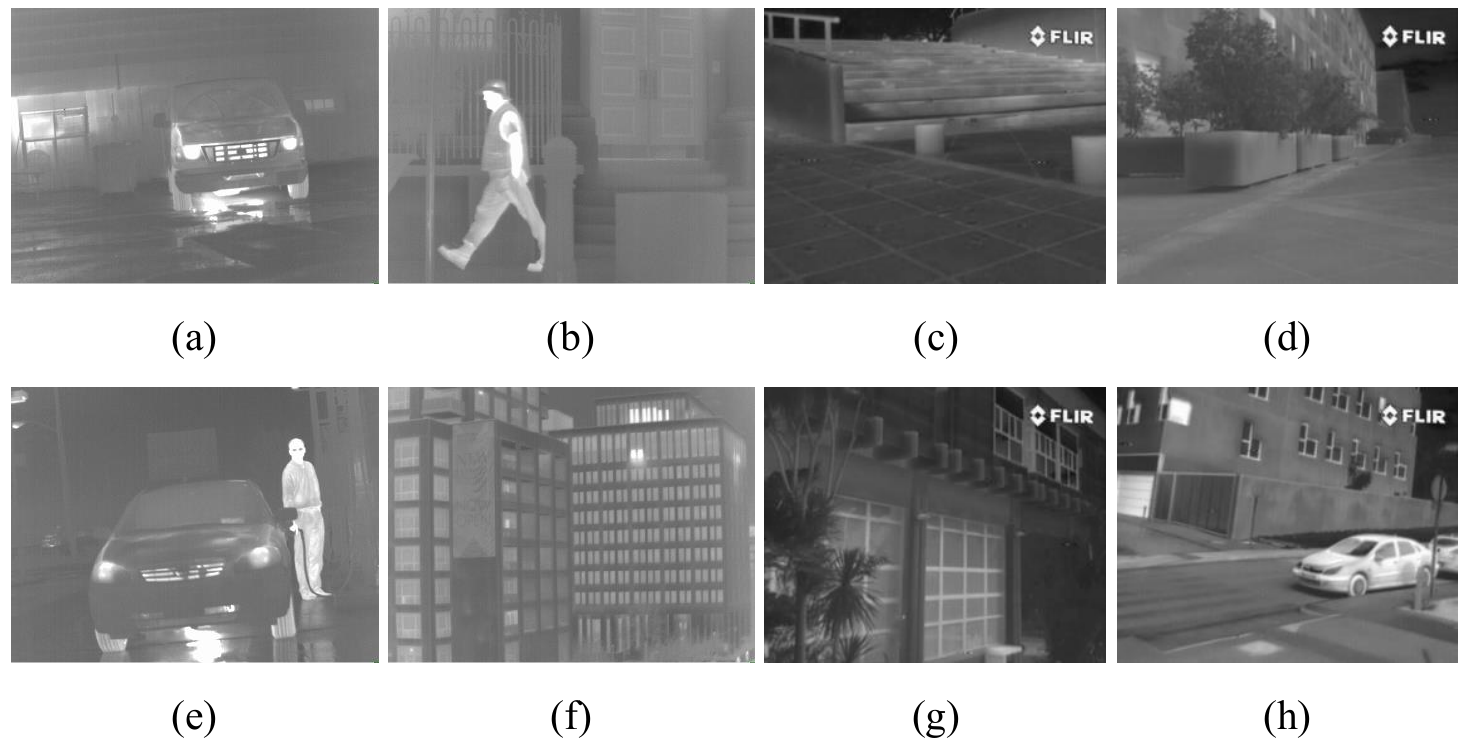}
\end{center}
\caption
{ \label{figure2}
  original images: (a) Truck, (b) Passerby, (c) Stairs, (d) Parterre, (e) Station, (f) Building, (g) Corridor, (h) Road.}
\end{figure}

To verify the performance of the proposed method, eight test images were employed from the infrared image databases of IRData (\url{http://www.dgp.toronto.edu/~nmorris/data/IRData/})  and CVC-15: Multimodal Stereo Dataset 2(\url{http://adas.cvc.uab.es/elektra/datasets/far-infra-red/}). These images are shown in Fig.\ref{figure2}. Our experiments were performed on a PC with an Intel CPU 2.8 GHz and 8 GB RAM using MATLAB R2014a. Four methods were adopted for comparison: ITV, ATV, L0TVPADMM, and OGSATVL1 .

For objective evaluation, we calculated the peak signal-to-noise ratio (PSNR), structural similarity (SSIM),  and relative error (RE). These are respectively defined as follows:
\begin{equation}\label{eq38}
\text {PSNR}(\bm X,\bm Y) = 10\text {log}_{10}\frac{{{255}^{2}}}{\frac{1}{{{N}^{2}}}\sum\limits_{i=1}^{N}{\sum\limits_{j=1}^{N}{{{({{X}_{ij}}-{{Y}_{ij}})}^{2}}}}},
\end{equation}
\begin{equation}\label{eq39}
\text{SSIM}(\bm X,\bm Y)=\frac{(2u_{\bm X}u_{\bm Y}+(255k_1)^2)(2\sigma_{\bm {XY}}+(255k_2)^2)}{(u_{\bm X}^2+u_{\bm Y}^2+(255k_1)^2)(\sigma_{\bm X}^2+\sigma_{\bm Y}^2+(255k_2)^2)},
\end{equation}
\begin{equation}\label{eq40}
\text {RE}(\bm X,\bm Y) = \frac{{{\left\| \bm Y-\bm X \right\|}_{2}}}{{{\left\| \bm X \right\|}_{2}}}.
\end{equation}

In general, larger values of PSNR and SSIM and smaller values of RE indicate better performance. In the experiment, we focus on the PSNR, while taking into account SSIM and RE. In all experiments, we set the experimental parameters as follows: $\beta_1 = 1$, $\beta_2 = 500$, $\beta_3 = 1$, and $\gamma = 1.618$. In addition, the blur kernel used in the experiment was generated by MATLAB built-in command ``fspecial (`gaussian', 7, 5)'', which gives a $7\times7$ Gaussian blur with a standard deviation of 5. The blur artifacts were also generated by the MATLAB command ``imfilter (Img, psf, `circular', `conv')'' under periodic boundary conditions, where ``Img'' represents the original image and ``psf'' represents the blur kernel.

\subsection{ Parameter optimization}
First, we determined a good value of K  for different images. In the experiment, we blurred the images Passerby, Station, and Truck with the blur kernel and corrupted them with 40\% salt-and-pepper noise. We set $p = 0.5$ and $p = 2/3$ to choose the best value of $K$.
\begin{figure}[htbp!]
\begin{center}
\includegraphics{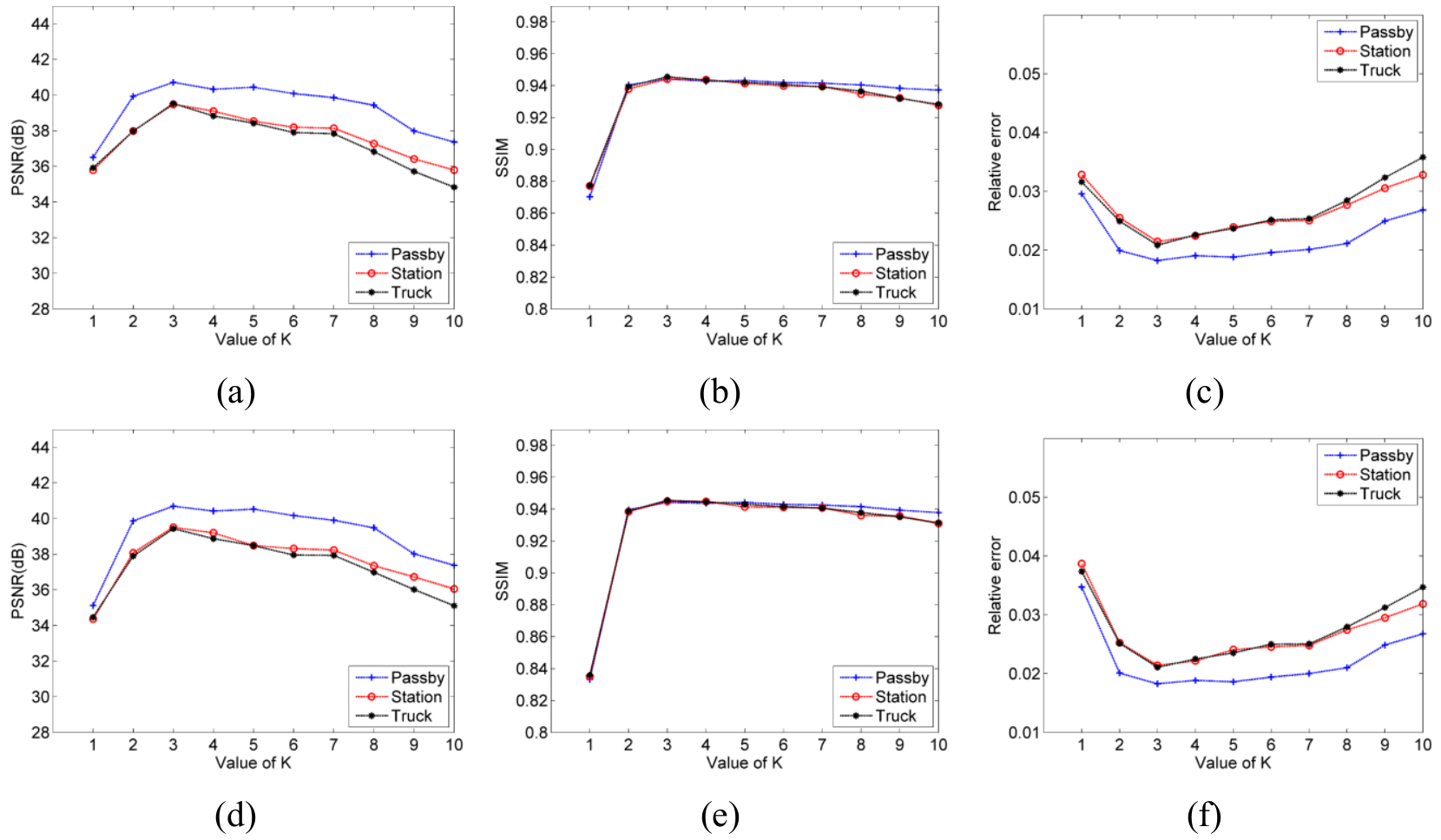}
\end{center}
\caption
{ \label{figure3}
  Results of our method for various values of group size K: (a) PSNR, (b) SSIM, and (c) RE for p = 0.5; (d)PSNR, (e) SSIM, and (f) RE for p = 2/3.}
\end{figure}

It can be seen from Fig.\ref{figure3} that for the three test images, the difference in the curves is not obvious for $p=0.5$ and $p=2/3$. Moreover, considering the three values of PSNR, SSIM, and RE, $K = 3$ is still the best parameter. Therefore, in all experiments, we consistently used this value.

Next, we evaluated the best value for regularization parameter $\mu$ for different images.  In this experiment, we blurred Passerby, Station, and Truck by the blur kernel and corrupted it with four different levels of salt-and-pepper noise. For each level of noise, we tested both $p=0.5$ and $p=2/3$.
\begin{figure}[htbp!]
\begin{center}
\includegraphics{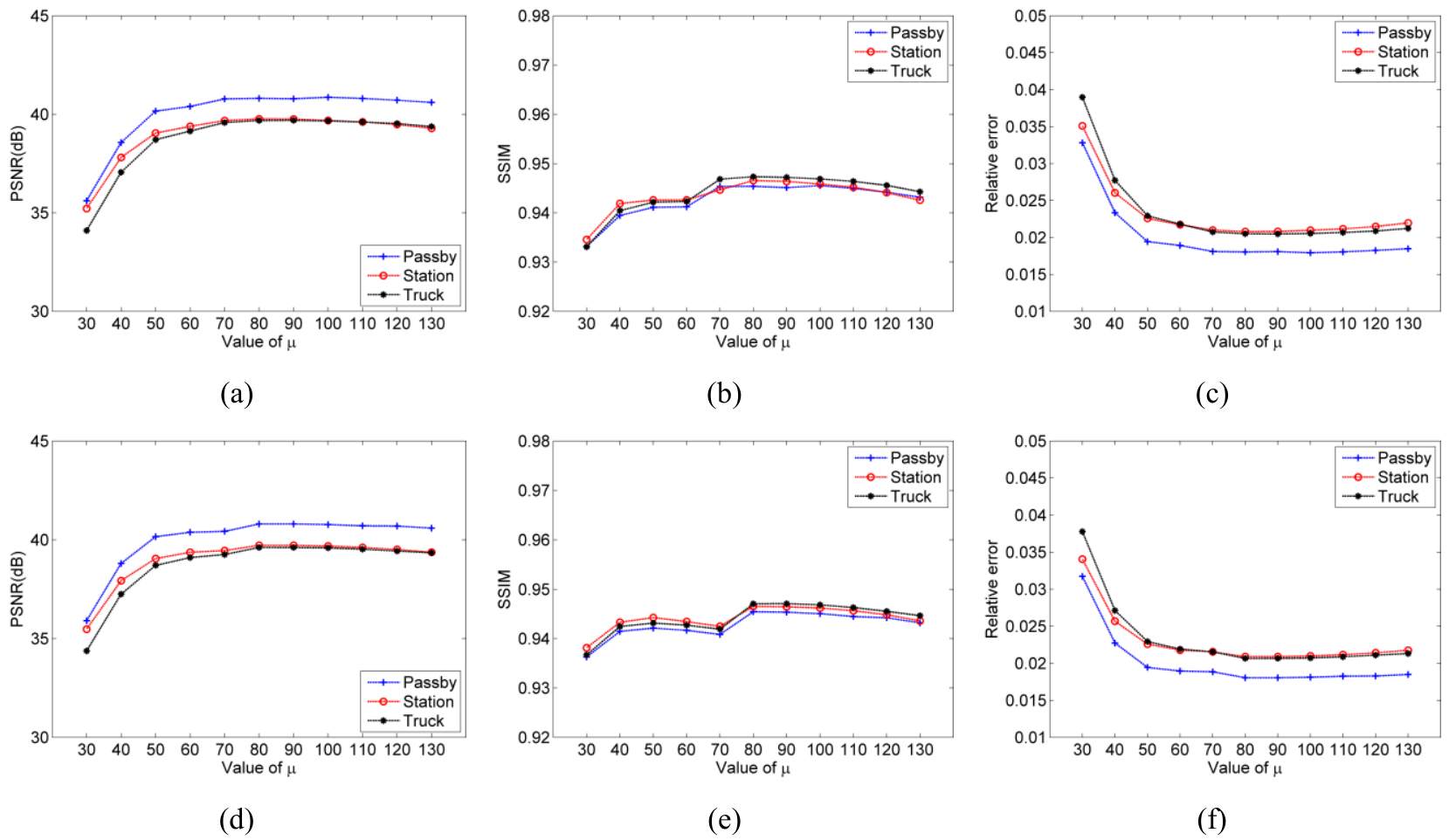}
\end{center}
\caption
{ \label{figure4}
  Results of our method for various values of regularization parameter $\mu$  when $\sigma = 30$: (a) PSNR, (b) SSIM, and (c) RE for $p = 0.5$; (d) PSNR, (e) SSIM, and (f) RE for $p = 2/3$.}
\end{figure}
\begin{figure}[htbp!]
\begin{center}
\includegraphics{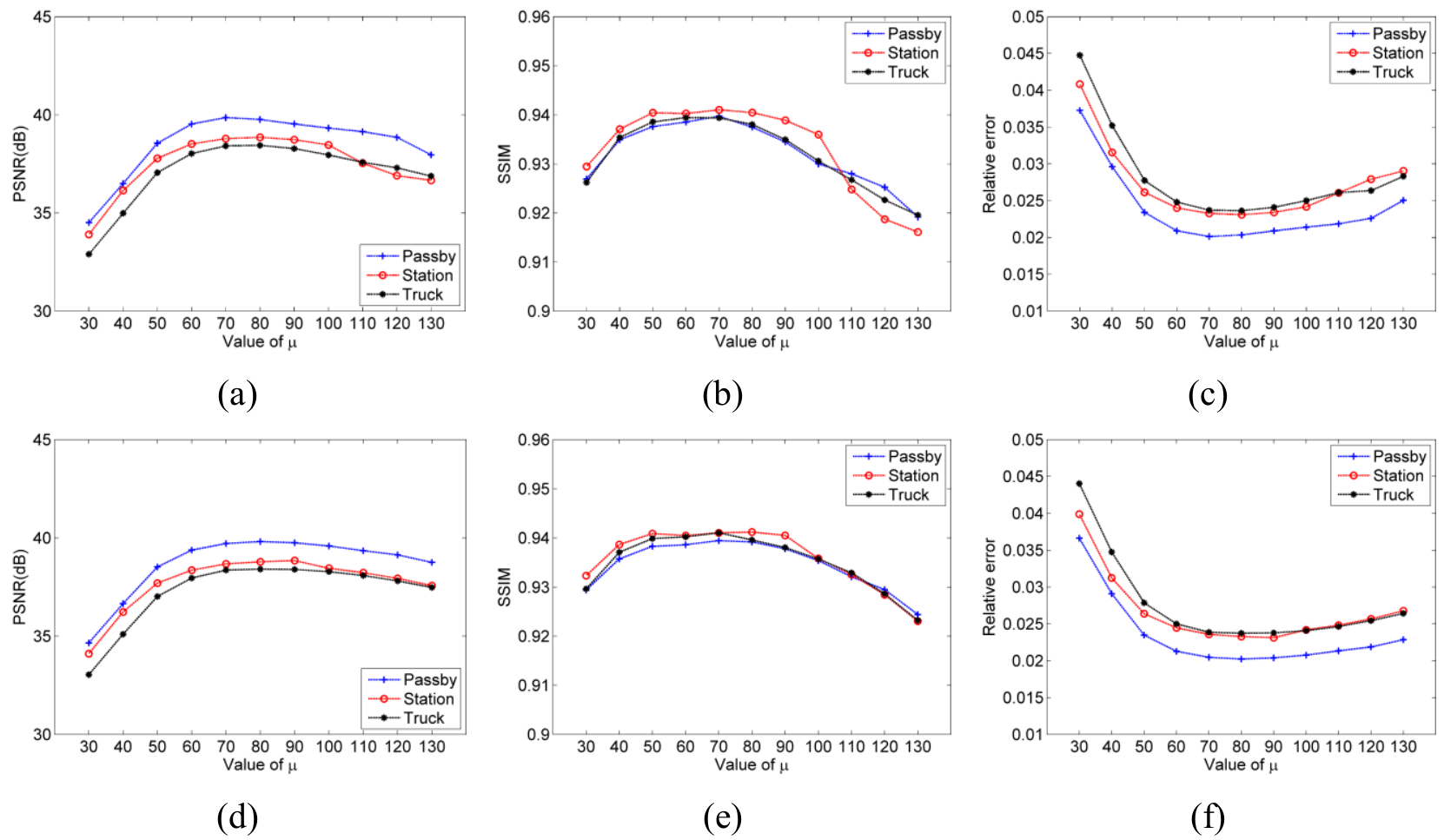}
\end{center}
\caption
{ \label{figure5}
  Results of our method for various values of regularization parameter $\mu$ when $\sigma = 40$: (a) PSNR, (b) SSIM, and (c) RE for $p = 0.5$; (d) PSNR, (e) SSIM, and (f) RE for $p = 2/3$.}
\end{figure}
\begin{figure}[htbp!]
\begin{center}
\includegraphics{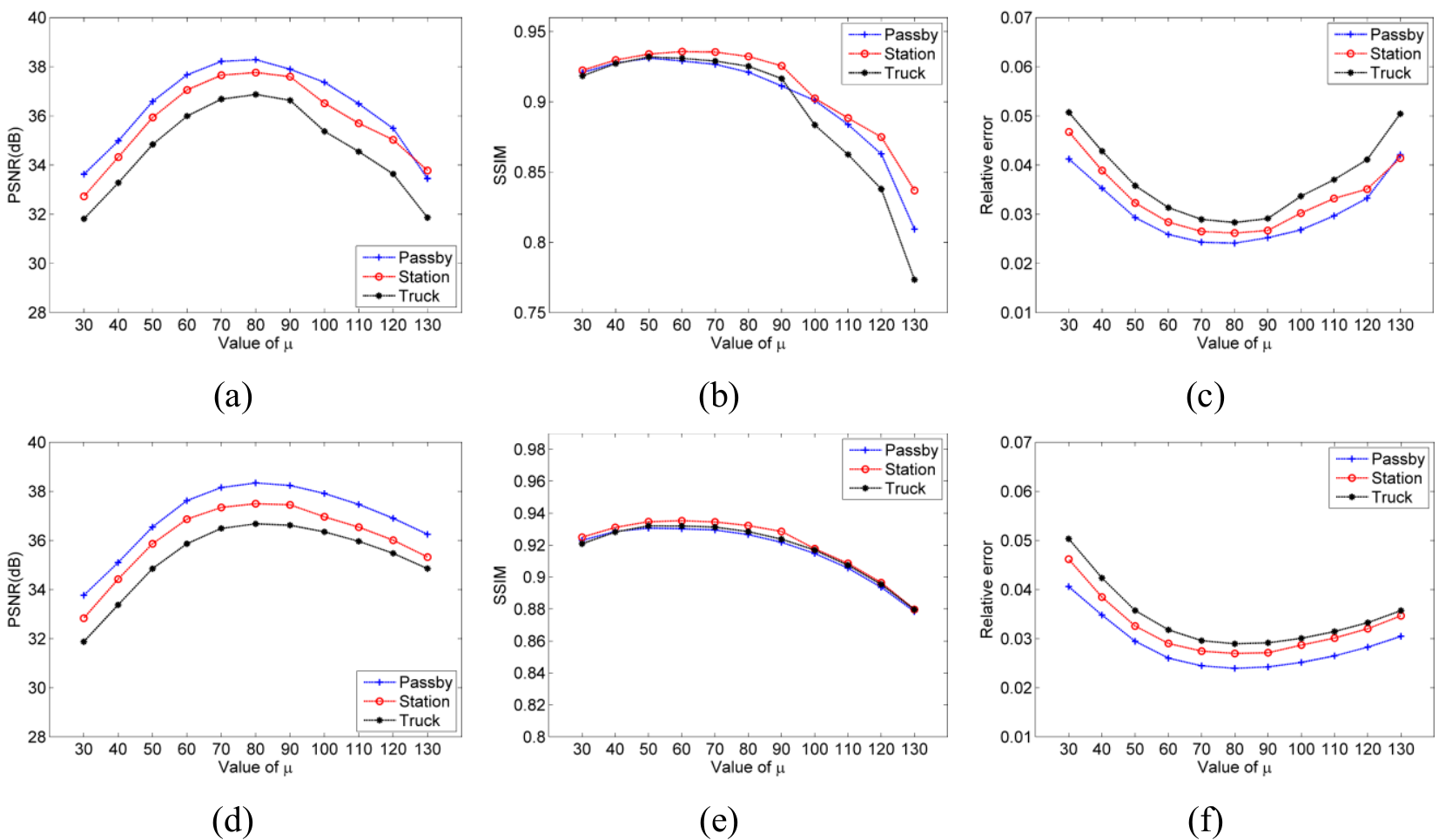}
\end{center}
\caption
{ \label{figure6}
  Results of our method for various values of regularization parameter $\mu$ when $\sigma = 50$: (a) PSNR, (b) SSIM, and (c) RE for $p = 0.5$; (d) PSNR, (e) SSIM, and (f) RE for $p = 2/3$.}
\end{figure}
\begin{figure}[htbp!]
\begin{center}
\includegraphics{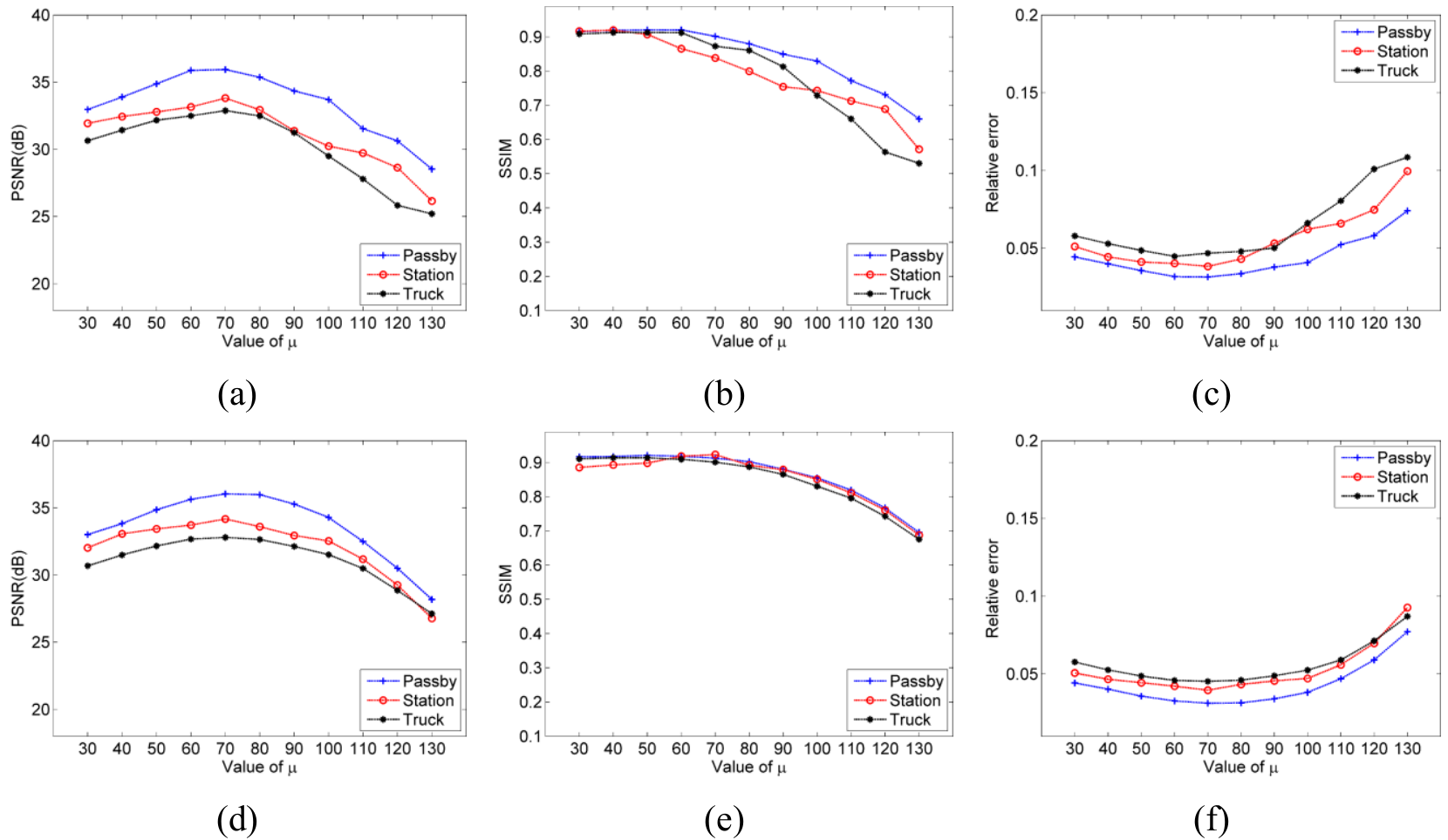}
\end{center}
\caption
{ \label{figure7}
  Results of our method for various values of regularization parameter $\mu$ when $\sigma = 60$: (a) PSNR, (b) SSIM, and (c) RE for $p = 0.5$; (d) PSNR, (e) SSIM, and (f) RE for $p = 2/3$.}
\end{figure}

Figures \ref{figure4} to \ref{figure7} show that for the three different images, the range of ¦Ì values changes basically for each of the two $p$ values. Therefore, in the subsequent experiments, for the four levels of noise, 30\%, 40\%, 50\%, and 60\%, the regularization parameter was set to 90, 80, 80, and 70, respectively.

\begin{figure}[htbp!]
\begin{center}
\includegraphics{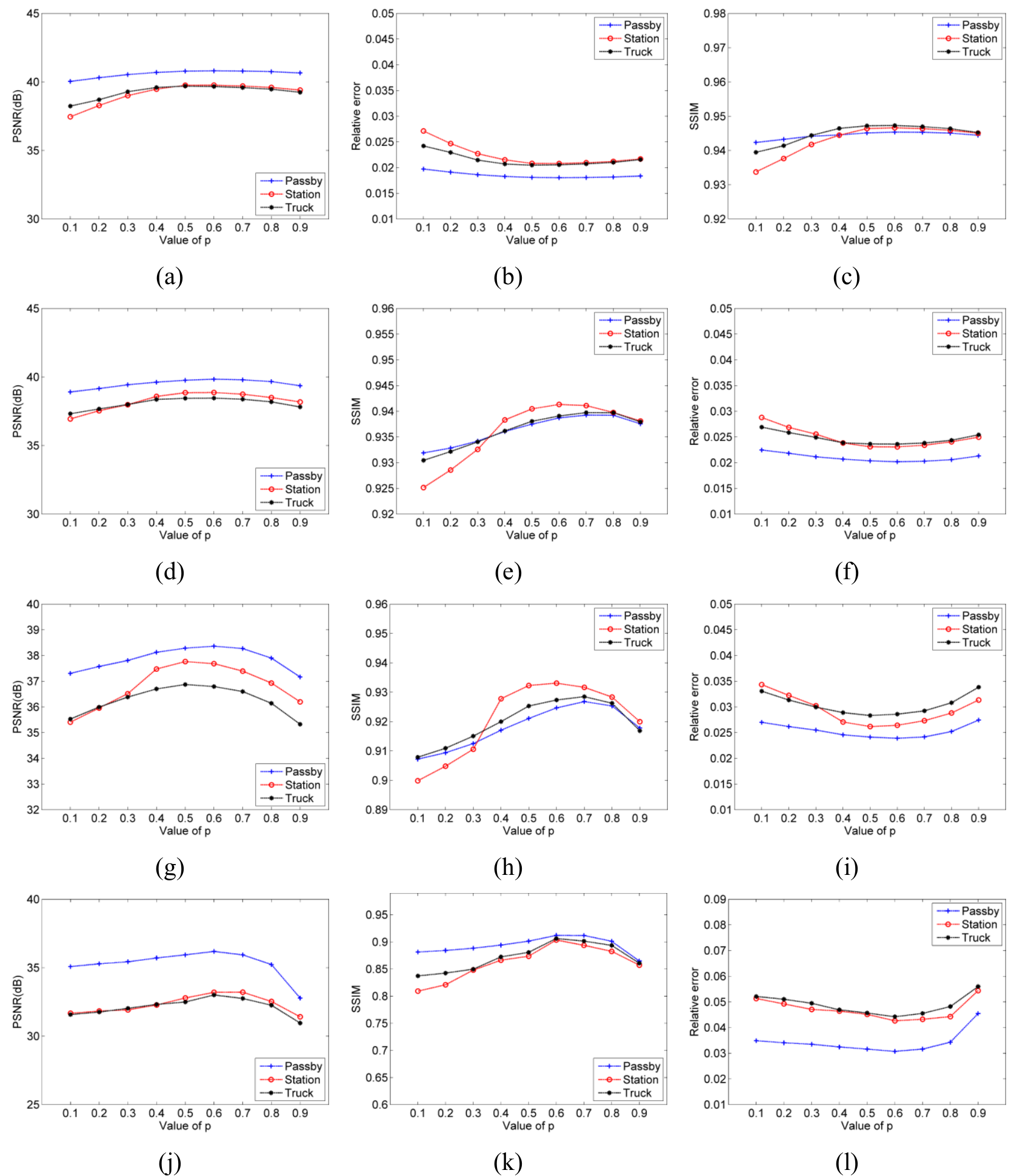}
\end{center}
\caption
{ \label{figure8}
 Results of our method for various values of parameter $p$  of $Lp$: (a) PSNR, (b) SSIM, and (c) RE for $\sigma=30$, $\mu=90$ ; (d) PSNR, (e) SSIM, and (f) RE for $\sigma=40$, $\mu=80$ ; (g) PSNR, (h) SSIM, and (i) RE for $\sigma=50$, $\mu=80$ ; (j) PSNR, (k) SSIM, and (l) RE for $\sigma=60$, $\mu=70$ .}
\end{figure}

In Figs.\ref{figure4} to \ref{figure7}, the curve is smoother at $p = 2/3$ under different noise levels, but the peak value are slightly lower than for $p = 0.5$. Hence, we also tested different values of parameter $p$. Figure \ref{figure8} shows that as the noise level increases, the performance metrics become more sensitive to $p$-value. Therefore, taking the behavior and absolute value of the results into account, we set the $p$ values as 0.5, 0.6, 0.6, and 0.6 for noise levels of 30\%, 40\%, 50\%, and 60\%, respectively.

\subsection{Comparison of OGSATVLp with Fast ADMM and ADMM}
To verify the algorithm converges if Fast ADMM is used, we tested the eight test images blurred with the blur kernel and corrupted with salt-and-pepper noise levels of 30\% and 40\%. The iteration stopping criterion was that the RE was less than 0.00001 or the Maximum inner iterations NIt was greater than 500.

\begin{figure}[htbp!]
\begin{center}
\includegraphics[width=12cm]{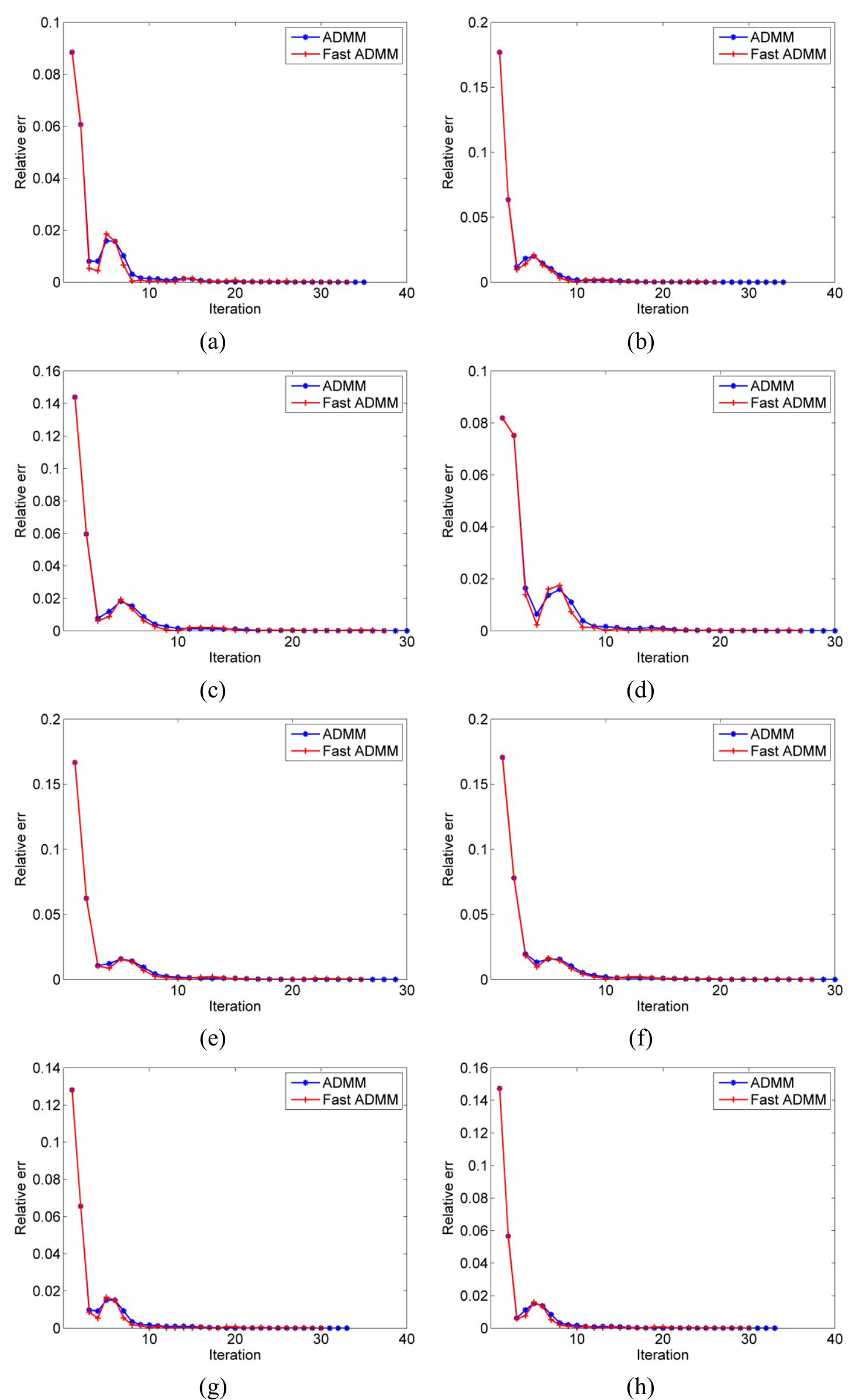}
\end{center}
\caption
{ \label{figure9}
 Comparison of OGSATVLp with Fast ADMM and ADMM when $\sigma=30$: RE for (a) Building, (b) Corridor, (c) Parterre, (d) Passerby, (e) Road, (f) Stairs, (g) Station, and (h) Truck.}
\end{figure}
\begin{figure}[htbp!]
\begin{center}
\includegraphics[width=12cm]{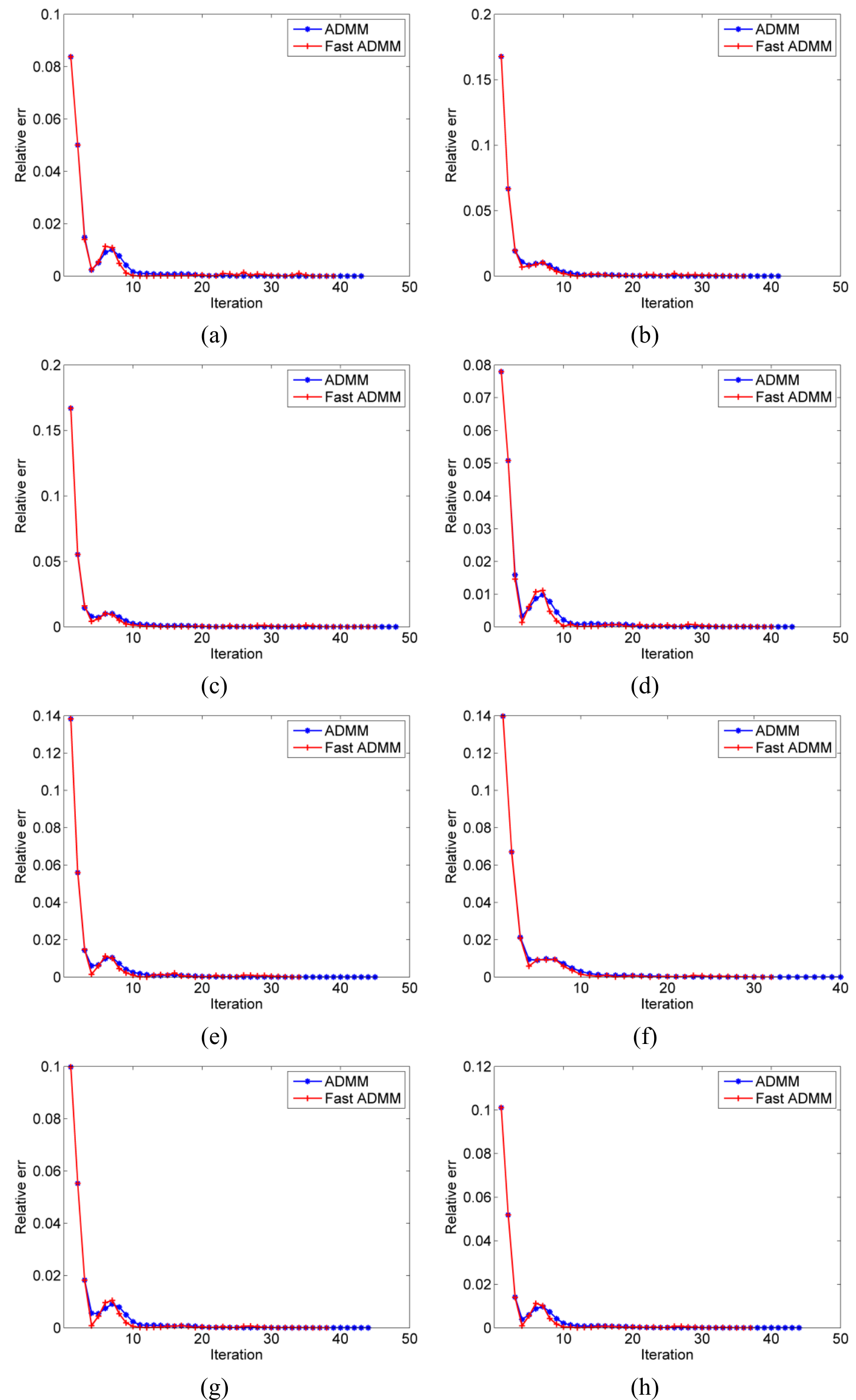}
\end{center}
\caption
{ \label{figure10}
 Comparison of OGSATVLp with Fast ADMM and ADMM when $\sigma=40$: RE for (a) Building, (b) Corridor, (c) Parterre, (d) Passerby, (e) Road, (f) Stairs, (g) Station, and (h) Truck.}
\end{figure}

The test results (Figs.\ref{figure9} and \ref{figure10}) show that as the noise level increases, ADMM increase the number of iterations of the eight images until the exit condition is reached. In contrast, the error curve of the Fast ADMM algorithm is faster and steeper, and the exit condition can be reached in fewer iterations. Moreover, for some images, the reduction in the number of iterations obtained by Fast ADMM is more pronounced at higher noise levels (Figs. \ref{figure9}(e)-(f) and \ref{figure10}(e)-(f)).

\subsection{Comparison with the OGSATVL1 algorithm}
In this section we compare the proposed method with the OGSATVL1 method. In the experiment, the salt-and-pepper noise levels of 30\% to 60\% were used to corrupt the six test images, which were then blurred with $7 \times7$ and $15\times15$ Gaussian blur kernels with a standard deviation of 5 (MATLAB commands ``fspecial(`gaussian', 7, 5)'' and ``fspecial(`gaussian', 15, 5), respectively)'' as well as a $7\times7$ mean blur kernel. The performances of both methods were then compared. For the OGSATVL1 method, to ensure that the maximum PSNR value could be obtained, the parameters of the method were individually set for each image. The parameters of the proposed method were set according to the above values.

First, for the $7\times7$ Gaussian blur kernel, the PSNR, SSIM, and RE numerical results of the test images are shown in Table \ref{table1}. The table shows that the results of the proposed method for all the test images at all noise levels are substantially better than those of the OGSATVL1 method, and this difference gradually increases as the noise level increases.

\begin{table}[htbp]
  \centering
  \caption{Results for the $7\times7$ Gaussian blur kernel}
  \footnotesize
    \begin{tabular}{lccccccccc}
    \toprule
    \multirow{2}[4]{*}{Images} & \multirow{2}[4]{*}{Noise level(dB)} & \multicolumn{4}{c}{OGSATVL1}  & \multicolumn{4}{c}{OGSATVLp} \\
\cmidrule{3-10}          &       & $\mu$     & PSNR(dB) & SSIM  & RE    & $\mu$     & PSNR(dB) & SSIM  & RE \\
    \midrule
    \multirow{4}[8]{*}{Passerby} & 30    & 80    & 40.4985  & 0.9437  & 0.0194  & 90    & \textbf{40.8630 } & \textbf{0.9455 } & \textbf{0.0179 } \\
\cmidrule{2-10}          & 40    & 80    & 38.8020  & 0.9331  & 0.0248  & 80    & \textbf{39.8587 } & \textbf{0.9397 } & \textbf{0.0201 } \\
\cmidrule{2-10}          & 50    & 60    & 36.2239  & 0.9182  & 0.0454  & 80    & \textbf{38.3621 } & \textbf{0.9247 } & \textbf{0.0239 } \\
\cmidrule{2-10}          & 60    & 30    & 32.7773  & 0.9186  & 0.2247  & 70    & \textbf{36.8660 } & \textbf{0.9285 } & \textbf{0.0307 } \\
    \midrule
    \multirow{4}[8]{*}{Station} & 30    & 80    & 39.2199  & 0.9440  & 0.0232  & 90    & \textbf{39.7670 } & \textbf{0.9466 } & \textbf{0.0208 } \\
\cmidrule{2-10}          & 40    & 80    & 37.6250  & 0.9335  & 0.0289  & 80    & \textbf{38.8663 } & \textbf{0.9413 } & \textbf{0.0231 } \\
\cmidrule{2-10}          & 50    & 60    & 35.2442  & 0.9207  & 0.0510  & 80    & \textbf{37.7629 } & \textbf{0.9323 } & \textbf{0.0262 } \\
\cmidrule{2-10}          & 60    & 30    & 31.5775  & 0.9166  & 0.2246  & 70    & \textbf{35.8792 } & \textbf{0.9320 } & \textbf{0.0412 } \\
    \midrule
    \multirow{4}[8]{*}{Truck} & 30    & 80    & 39.0780  & 0.9443  & 0.0229  & 90    & \textbf{39.6943 } & \textbf{0.9472 } & \textbf{0.0205 } \\
\cmidrule{2-10}          & 40    & 80    & 37.1923  & 0.9330  & 0.0289  & 80    & \textbf{38.4550 } & \textbf{0.9391 } & \textbf{0.0236 } \\
\cmidrule{2-10}          & 50    & 70    & 34.3116  & 0.9090  & 0.0518  & 80    & \textbf{36.8715 } & \textbf{0.9253 } & \textbf{0.0283 } \\
\cmidrule{2-10}          & 60    & 40    & 29.9586  & 0.9119  & 0.2173  & 70    & \textbf{34.7389 } & \textbf{0.9287 } & \textbf{0.0442 } \\
    \midrule
    \multirow{4}[8]{*}{Parterre} & 30    & 80    & 42.2099  & 0.9790  & 0.0247  & 90    & \textbf{43.0244 } & \textbf{0.9804 } & \textbf{0.0225 } \\
\cmidrule{2-10}          & 40    & 60    & 41.4431  & 0.9761  & 0.0322  & 80    & \textbf{42.5242 } & \textbf{0.9781 } & \textbf{0.0246 } \\
\cmidrule{2-10}          & 50    & 40    & 39.1755  & 0.9688  & 0.0617  & 80    & \textbf{41.2265 } & \textbf{0.9729 } & \textbf{0.0267 } \\
\cmidrule{2-10}          & 60    & 30    & 36.4261  & 0.9545  & 0.2483  & 70    & \textbf{39.5382 } & \textbf{0.9638 } & \textbf{0.0302 } \\
    \midrule
    \multirow{4}[8]{*}{Stairs} & 30    & 80    & 42.1100  & 0.9804  & 0.0307  & 90    & \textbf{43.0574 } & \textbf{0.9826 } & \textbf{0.0279 } \\
\cmidrule{2-10}          & 40    & 60    & 40.8109  & 0.9773  & 0.0408  & 80    & \textbf{42.1333 } & \textbf{0.9801 } & \textbf{0.0304 } \\
\cmidrule{2-10}          & 50    & 40    & 39.0959  & 0.9696  & 0.0751  & 80    & \textbf{40.6847 } & \textbf{0.9709 } & \textbf{0.0347 } \\
\cmidrule{2-10}          & 60    & 30    & 36.0909  & 0.9565  & 0.2976  & 70    & \textbf{38.3841 } & \textbf{0.9622 } & \textbf{0.0403 } \\
    \midrule
    \multirow{4}[8]{*}{Corridor} & 30    & 70    & 41.3814  & 0.9787  & 0.0347  & 90    & \textbf{42.3201 } & \textbf{0.9813 } & \textbf{0.0304 } \\
\cmidrule{2-10}          & 40    & 60    & 39.9433  & 0.9739  & 0.0433  & 80    & \textbf{41.2782 } & \textbf{0.9775 } & \textbf{0.0331 } \\
\cmidrule{2-10}          & 50    & 40    & 38.5279  & 0.9649  & 0.0904  & 80    & \textbf{40.2169 } & \textbf{0.9719 } & \textbf{0.0368 } \\
\cmidrule{2-10}          & 60    & 30    & 35.3223  & 0.9478  & 0.2944  & 70    & \textbf{37.9723 } & \textbf{0.9562 } & \textbf{0.0439 } \\
    \bottomrule
    \end{tabular}%
  \label{table1}%
\end{table}%

\begin{figure}[htbp!]
\begin{center}
\includegraphics{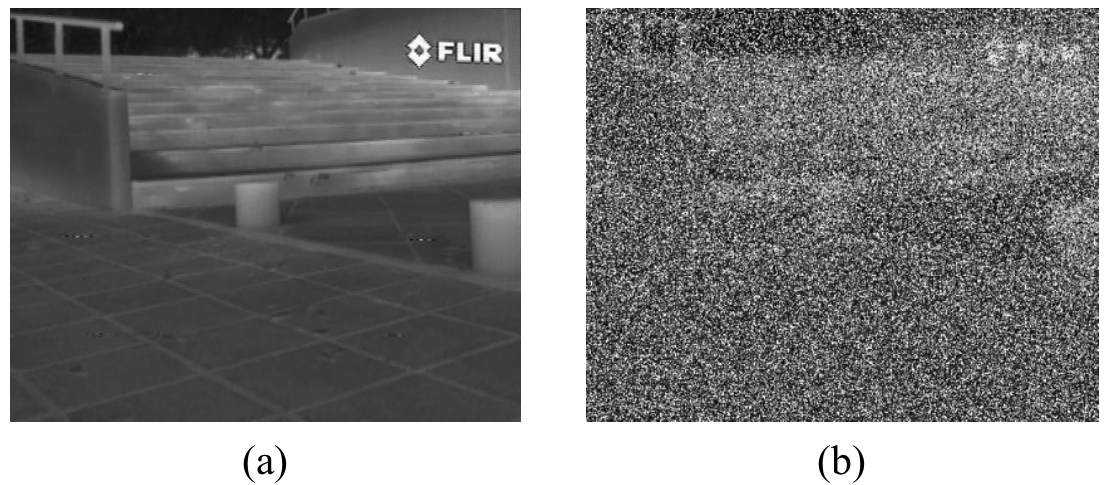}
\end{center}
\caption
{ \label{figure11}
 Stairs image: (a) blurred with a $7 \times 7$ Gaussian blur kernel and (b) corrupted with 50\% salt-and-pepper noise.}
\end{figure}

\begin{figure}[htbp!]
\begin{center}
\includegraphics{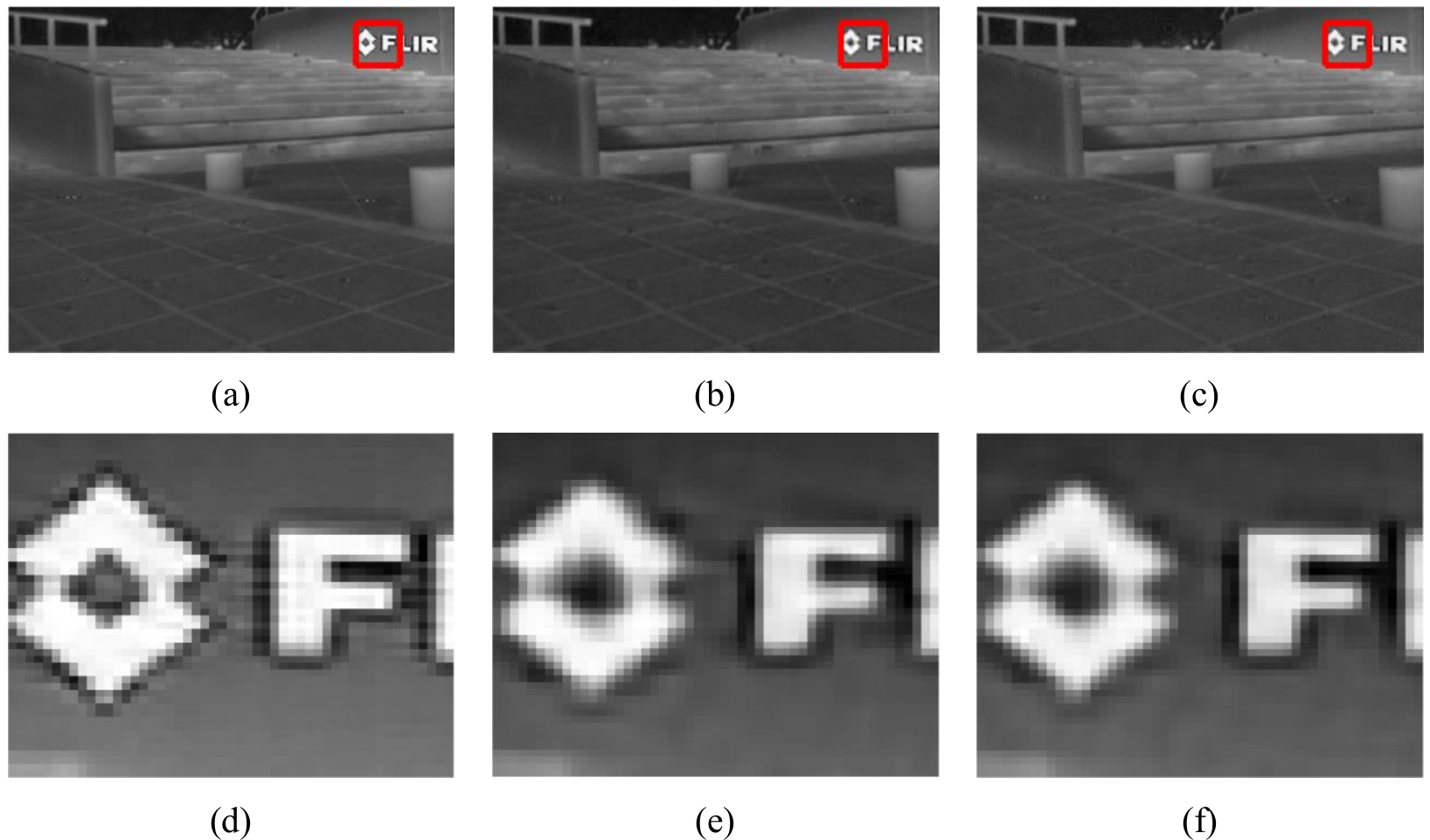}
\end{center}
\caption
{ \label{figure12}
 Results for the Stairs image in Fig.\ref{figure11}: (a) original image, (b) results for OGSATVL1, and (c) results for  OGSATVLp. Enlarged view of the area enclosed by the red squares: (d) original image, (e) results for OGSATVL1, and (f) results for  OGSATVLp.}
\end{figure}

Next, we present a visual comparison of the results. We blurred the Stairs image with a $7 \times 7 $ Gaussian blur kernel with a standard deviation of 5 and added 50\% salt-and-pepper noise (Fig.\ref{figure11}). Figure \ref{figure12} shows the results obtained by OGSATVL1 and the proposed OGSATVLp method.
As can be seen from the enlarged images in the bottom row of Fig.\ref{figure12}, for image edges such as the font edges, the OGSATVL1 method does not obtain sufficiently clear results, there are significant block artifacts, and the corner points are too smooth. The boundaries obtained by the proposed OGSATVLp method are clearer, the block artifacts are less strong, and the corner points are closer to those of the original image.

In addition, the PSNR, SSIM, and RE numerical results of all six test images for these conditions are shown in Table \ref{table2} . The results show that the PSNR and RE values of the proposed method are significantly better than those of the OGSATVL1 method for all six test images, but the SSIM values of the OGSATVL1 results are better than the proposed OGSATVLp method for some test images. Considering that the maximum PSNR value is the main criterion for the parameter selection, the SSIM value is only used here for reference.

\begin{table}[htbp]
  \centering
  \caption{Results for $15\times15$ Gaussian blur kernel}
   \footnotesize
    \begin{tabular}{lccrrrcrrr}
        \toprule
    \multirow{2}[4]{*}{Images} & \multirow{2}[4]{*}{Noise level(\%)} & \multicolumn{4}{c}{OGSATVL1}  & \multicolumn{4}{c}{OGSATVLp} \\
\cmidrule{3-10}          &       & $\mu$     & PSNR(dB) & SSIM  & RE    & $\mu$     & PSNR(dB) & SSIM  & RE \\
    \midrule
    \multirow{4}[8]{*}{Passerby} & 30    & 120   & 32.9436 & 0.8997 & 0.0456 & 270   & \textbf{34.6057} & \textbf{0.9003} & \textbf{0.0368} \\
\cmidrule{2-10}          & 40    & 150   & 32.4779 & \textbf{0.8963} & 0.0513 & 220   & \textbf{32.7835} & 0.8766 & \textbf{0.0454} \\
\cmidrule{2-10}          & 50    & 170   & 32.1993 & 0.8899 & 0.0537 & 190   & \textbf{32.9063} & \textbf{0.8940} & \textbf{0.0448} \\
\cmidrule{2-10}          & 60    & 150   & 31.3738 & 0.8800 & 0.0631 & 170   & \textbf{31.7850} & \textbf{0.8844} & \textbf{0.0510} \\
    \midrule
    \multirow{4}[8]{*}{Station} & 30    & 180   & 31.7350 & \textbf{0.8896} & 0.0533 & 220   & \textbf{32.4409} & 0.8754 & \textbf{0.0483} \\
\cmidrule{2-10}          & 40    & 170   & 30.9524 & 0.8840 & 0.0590 & 200   & \textbf{31.3053} & \textbf{0.8873} & \textbf{0.0551} \\
\cmidrule{2-10}          & 50    & 170   & 30.5016 & 0.8773 & 0.0644 & 170   & \textbf{31.5787} & \textbf{0.8800} & \textbf{0.0533} \\
\cmidrule{2-10}          & 60    & 150   & 29.7653 & \textbf{0.8682} & 0.0745 & 230   & \textbf{30.5709} & 0.8275 & \textbf{0.0599} \\
    \midrule
    \multirow{4}[8]{*}{Truck} & 30    & 240   & 31.3215 & 0.8894 & 0.0529 & 260   & \textbf{32.2856} & \textbf{0.8933} & \textbf{0.0480} \\
\cmidrule{2-10}          & 40    & 220   & 30.5913 & \textbf{0.8798} & 0.0599 & 180   & \textbf{31.6742} & 0.8569 & \textbf{0.0515} \\
\cmidrule{2-10}          & 50    & 170   & 29.9966 & \textbf{0.8769} & 0.0671 & 180   & \textbf{30.8385} & 0.8419 & \textbf{0.0567} \\
\cmidrule{2-10}          & 60    & 160   & 29.5562 & 0.8658 & 0.0730 & 190   & \textbf{30.1136} & \textbf{0.8684} & \textbf{0.0617} \\
    \midrule
    \multirow{4}[8]{*}{Parterre} & 30    & 90    & 35.0781 & \textbf{0.9191} & 0.0479 & 210   & \textbf{35.1603} & 0.9099 & \textbf{0.0445} \\
\cmidrule{2-10}          & 40    & 100   & 34.9144 & \textbf{0.9186} & 0.0512 & 130   & \textbf{35.0604} & 0.9185 & \textbf{0.0450} \\
\cmidrule{2-10}          & 50    & 130   & 34.6906 & 0.9148 & 0.0572 & 130   & \textbf{35.1331} & \textbf{0.9177} & \textbf{0.0446} \\
\cmidrule{2-10}          & 60    & 110   & 34.2452 & \textbf{0.9120} & 0.0688 & 150   & \textbf{34.8958} & 0.9109 & \textbf{0.0459} \\
    \midrule
    \multirow{4}[8]{*}{Stairs} & 30    & 110   & 35.0344 & 0.9250 & 0.0580 & 190   & \textbf{36.2397} & \textbf{0.9257} & \textbf{0.0474} \\
\cmidrule{2-10}          & 40    & 120   & 34.7010 & \textbf{0.9228} & 0.0638 & 130   & \textbf{35.5195} & 0.9204 & \textbf{0.0515} \\
\cmidrule{2-10}          & 50    & 110   & 34.5878 & \textbf{0.9210} & 0.0703 & 130   & \textbf{35.2472} & 0.9175 & \textbf{0.0532} \\
\cmidrule{2-10}          & 60    & 120   & 33.9054 & 0.9132 & 0.0882 & 140   & \textbf{34.5833} & \textbf{0.9180} & \textbf{0.0574} \\
    \midrule
    \multirow{4}[8]{*}{Corridor} & 30    & 90    & 33.7068 & 0.9051 & 0.0668 & 100   & \textbf{34.9438} & \textbf{0.9210} & \textbf{0.0560} \\
\cmidrule{2-10}          & 40    & 140   & 33.5111 & 0.9016 & 0.0708 & 140   & \textbf{33.6087} & \textbf{0.9026} & \textbf{0.0653} \\
\cmidrule{2-10}          & 50    & 130   & 33.1528 & 0.8979 & 0.0819 & 140   & \textbf{33.6661} & \textbf{0.9039} & \textbf{0.0649} \\
\cmidrule{2-10}          & 60    & 120   & 32.5816 & 0.8885 & 0.1002 & 130   & \textbf{33.2624} & \textbf{0.8979} & \textbf{0.0679} \\
    \bottomrule
    \end{tabular}%
  \label{table2}%
\end{table}%

\begin{figure}[htbp!]
\begin{center}
\includegraphics{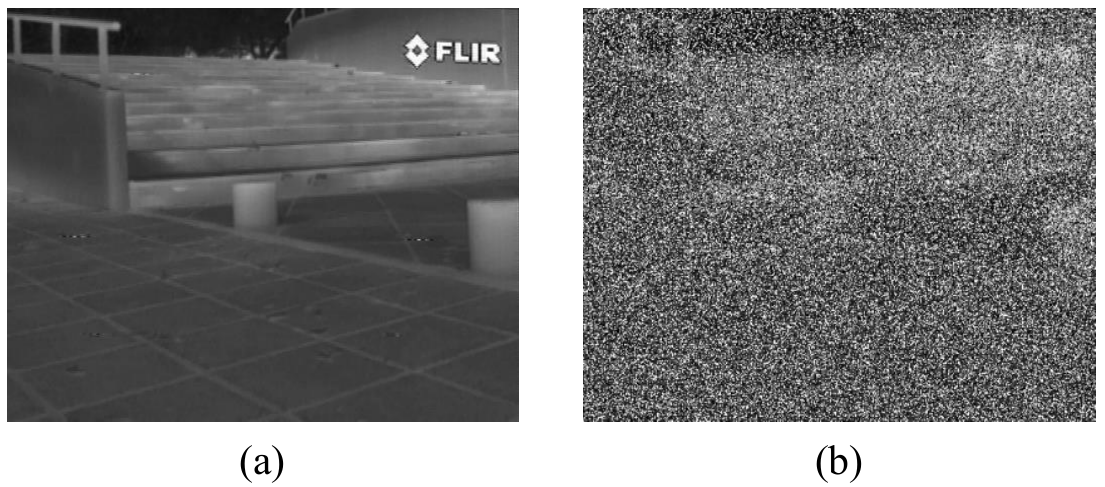}
\end{center}
\caption
{ \label{figure13}
 Stairs image: (a) blurred with a $15 \times 15$ Gaussian blur kernel and (b) corrupted with 50\% salt-and-pepper noise.}
\end{figure}

We compare the visual effects of the two methods on the test image Stairs for a $15 \times 15$ Gaussian blur kernel and with 50\% salt-and-pepper noise (as shown in Fig.\ref{figure13}). The magnified results show that the two methods have similar results on the image under Gaussian blur. In contrast, the boundary obtained by the proposed method is slightly clearer and the block artifacts are relatively faint.
\begin{figure}[htbp!]
\begin{center}
\includegraphics{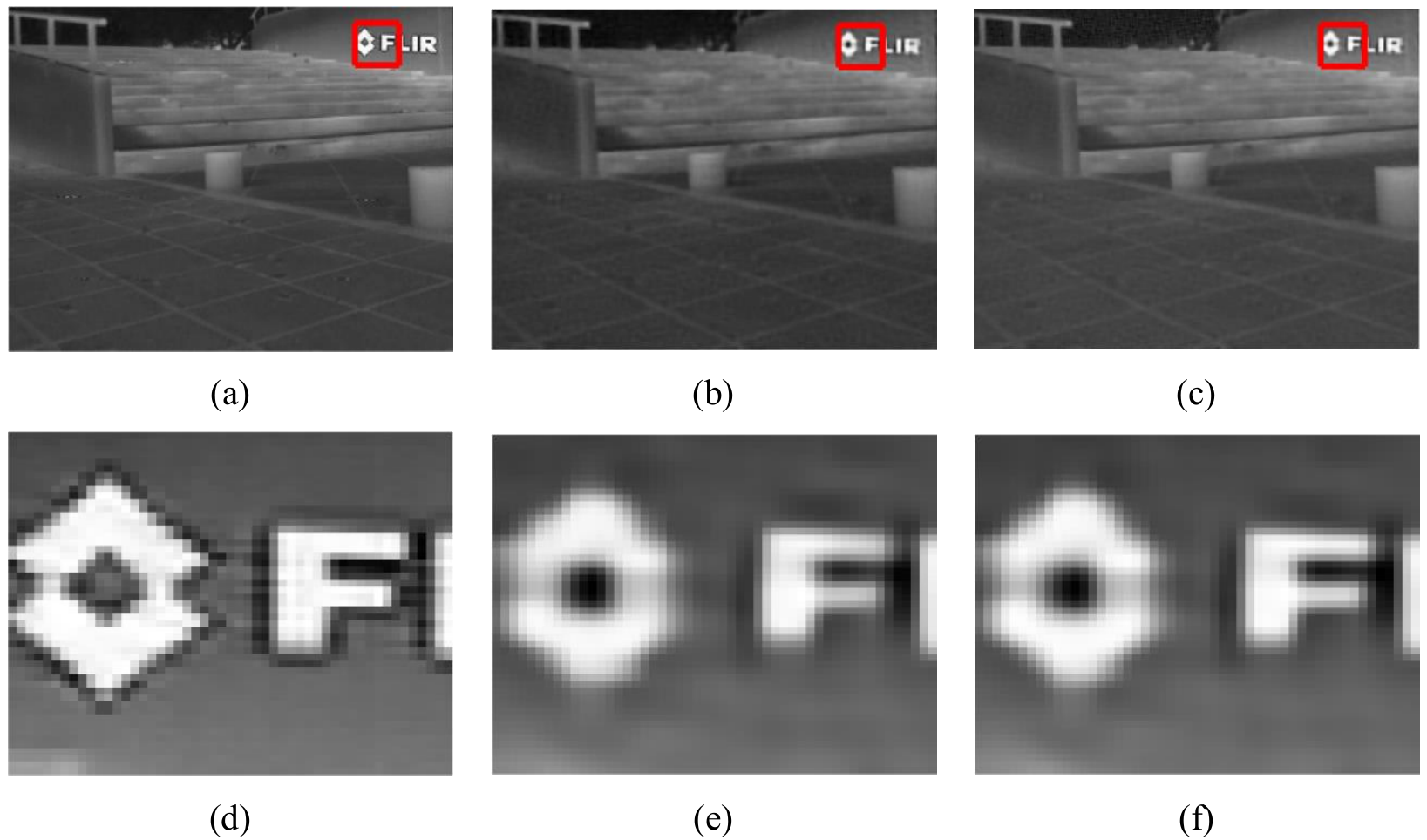}
\end{center}
\caption
{ \label{figure14}
 Results for the Stairs image in Fig.\ref{figure13}: (a) original image, (b) results for OGSATVL1, and (c) results for  OGSATVLp. Enlarged view of the area enclosed by the red squares: (d) original image, (e) results for OGSATVL1, and (f) results for  OGSATVLp.}
\end{figure}

Finally, the proposed method and OGSATVL1 are compared for images corrupted with a $7 \times 7$ mean blur kernel and  noise levels of 30\% to 60\%.  Table \ref{table3} shows that the results of this method are better than those of OGSATVL1 for all six test images under different noise levels. Moreover, the gap between the two results becomes significantly larger as the noise level increases.
\begin{table}[htbp]
  \centering
  \caption{ Results for the $7 \times 7$ mean blur kernel}
  \footnotesize
    \begin{tabular}{lccccccccc}
    \toprule
    \multirow{2}[4]{*}{Images} & \multirow{2}[4]{*}{Noise level(\%)} & \multicolumn{4}{c}{OGSATVL1}  & \multicolumn{4}{c}{OGSATVLp} \\
\cmidrule{3-10}          &       & ¦Ì     & PSNR(dB) & SSIM  & RE    & ¦Ì     & PSNR(dB) & SSIM  & RE \\
    \midrule
    \multirow{4}[8]{*}{Passby} & 30    & 80    & 40.3906  & 0.9433  & 0.0290  & 90    & \textbf{40.8217 } & \textbf{0.9457 } & \textbf{0.0180 } \\
\cmidrule{2-10}          & 40    & 70    & 38.5239  & 0.9309  & 0.0709  & 80    & \textbf{40.1691 } & \textbf{0.9402 } & \textbf{0.0194 } \\
\cmidrule{2-10}          & 50    & 60    & 35.4876  & 0.9116  & 0.0359  & 70    & \textbf{38.7155 } & \textbf{0.9303 } & \textbf{0.0229 } \\
\cmidrule{2-10}          & 60    & 40    & 32.6870  & 0.8923  & 0.1145  & 60    & \textbf{36.0681 } & \textbf{0.9135 } & \textbf{0.0311 } \\
    \midrule
    \multirow{4}[8]{*}{Station} & 30    & 80    & 39.2587  & 0.9444  & 0.0329  & 80    & \textbf{39.8743 } & \textbf{0.9474 } & \textbf{0.0205 } \\
\cmidrule{2-10}          & 40    & 70    & 37.5000  & 0.9336  & 0.0676  & 70    & \textbf{39.1676 } & \textbf{0.9415 } & \textbf{0.0223 } \\
\cmidrule{2-10}          & 50    & 60    & 35.2170  & 0.9158  & 0.0396  & 70    & \textbf{37.9245 } & \textbf{0.9348 } & \textbf{0.0257 } \\
\cmidrule{2-10}          & 60    & 30    & 31.6639  & 0.9063  & 0.1217  & 60    & \textbf{35.6368 } & \textbf{0.9195 } & \textbf{0.0334 } \\
    \midrule
    \multirow{4}[8]{*}{Truck} & 30    & 80    & 39.2287  & 0.9437  & 0.0327  & 90    & \textbf{39.9195 } & \textbf{0.9480 } & \textbf{0.0199 } \\
\cmidrule{2-10}          & 40    & 70    & 37.6325  & 0.9354  & 0.0661  & 70    & \textbf{39.1053 } & \textbf{0.9415 } & \textbf{0.0219 } \\
\cmidrule{2-10}          & 50    & 60    & 34.3421  & 0.9127  & 0.0406  & 70    & \textbf{37.4036 } & \textbf{0.9304 } & \textbf{0.0266 } \\
\cmidrule{2-10}          & 60    & 40    & 30.9083  & 0.8906  & 0.1171  & 60    & \textbf{34.6196 } & \textbf{0.9088 } & \textbf{0.0367 } \\
    \midrule
    \multirow{4}[8]{*}{parterre} & 30    & 80    & 40.6783  & 0.9583  & 0.0234  & 80    & \textbf{41.1108 } & \textbf{0.9606 } & \textbf{0.0224 } \\
\cmidrule{2-10}          & 40    & 60    & 39.4515  & 0.9518  & 0.0282  & 70    & \textbf{40.3230 } & \textbf{0.9556 } & \textbf{0.0246 } \\
\cmidrule{2-10}          & 50    & 40    & 38.1268  & 0.9470  & 0.0421  & 50    & \textbf{39.5214 } & \textbf{0.9532 } & \textbf{0.0269 } \\
\cmidrule{2-10}          & 60    & 30    & 36.6059  & 0.9380  & 0.1361  & 50    & \textbf{38.5079 } & \textbf{0.9453 } & \textbf{0.0303 } \\
    \midrule
    \multirow{4}[8]{*}{Stairs} & 30    & 80    & 40.4054  & 0.9598  & 0.0289  & 80    & \textbf{40.9179 } & \textbf{0.9618 } & \textbf{0.0277 } \\
\cmidrule{2-10}          & 40    & 60    & 39.0759  & 0.9531  & 0.0355  & 70    & \textbf{40.0870 } & \textbf{0.9570 } & \textbf{0.0305 } \\
\cmidrule{2-10}          & 50    & 40    & 37.7394  & 0.9481  & 0.0537  & 50    & \textbf{39.1411 } & \textbf{0.9556 } & \textbf{0.0340 } \\
\cmidrule{2-10}          & 60    & 30    & 35.3994  & 0.9367  & 0.1802  & 40    & \textbf{37.7778 } & \textbf{0.9499 } & \textbf{0.0397 } \\
    \midrule
    \multirow{4}[8]{*}{corridor} & 30    & 70    & 39.7015  & 0.9598  & 0.0323  & 90    & \textbf{40.5221 } & \textbf{0.9639 } & \textbf{0.0295 } \\
\cmidrule{2-10}          & 40    & 60    & 38.5194  & 0.9517  & 0.0384  & 70    & \textbf{39.3766 } & \textbf{0.9547 } & \textbf{0.0336 } \\
\cmidrule{2-10}          & 50    & 40    & 37.0172  & 0.9445  & 0.0636  & 50    & \textbf{38.6434 } & \textbf{0.9535 } & \textbf{0.0366 } \\
\cmidrule{2-10}          & 60    & 30    & 34.3520  & 0.9274  & 0.1839  & 40    & \textbf{37.0657 } & \textbf{0.9454 } & \textbf{0.0439 } \\
    \bottomrule
    \end{tabular}%
  \label{table3}%
\end{table}%

We employ Stairs for the visual comparison for the mean kernel blur  and the results are shown in Fig.\ref{figure16}. The enlarged results show that there are a significant block artifacts at the edges after OGSATVL1, and there is also a small block artifact in the partially smooth region. In addition, because of excessive smoothing, the restored image as a whole is blurred. In contrast, the results of the proposed method are relatively clear, the block artifacts are fainter, and the overall recovered results are better.
\begin{figure}[htbp!]
\begin{center}
\includegraphics{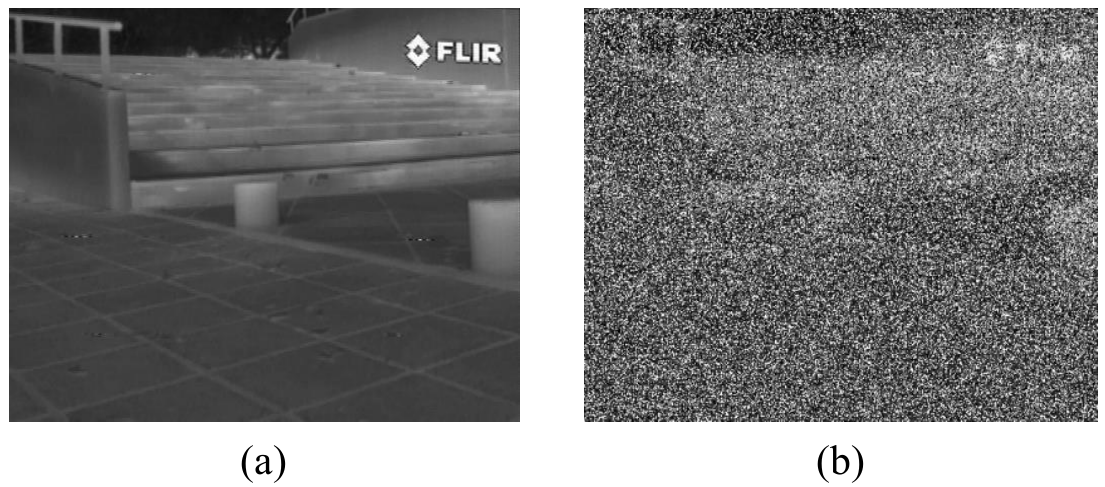}
\end{center}
\caption
{ \label{figure15}
 Stairs image: (a) blurred with a $7 \times 7$ mean blur kernel and (b) corrupted with 50\% salt-and-pepper noise.}
\end{figure}
\begin{figure}[htbp!]
\begin{center}
\includegraphics{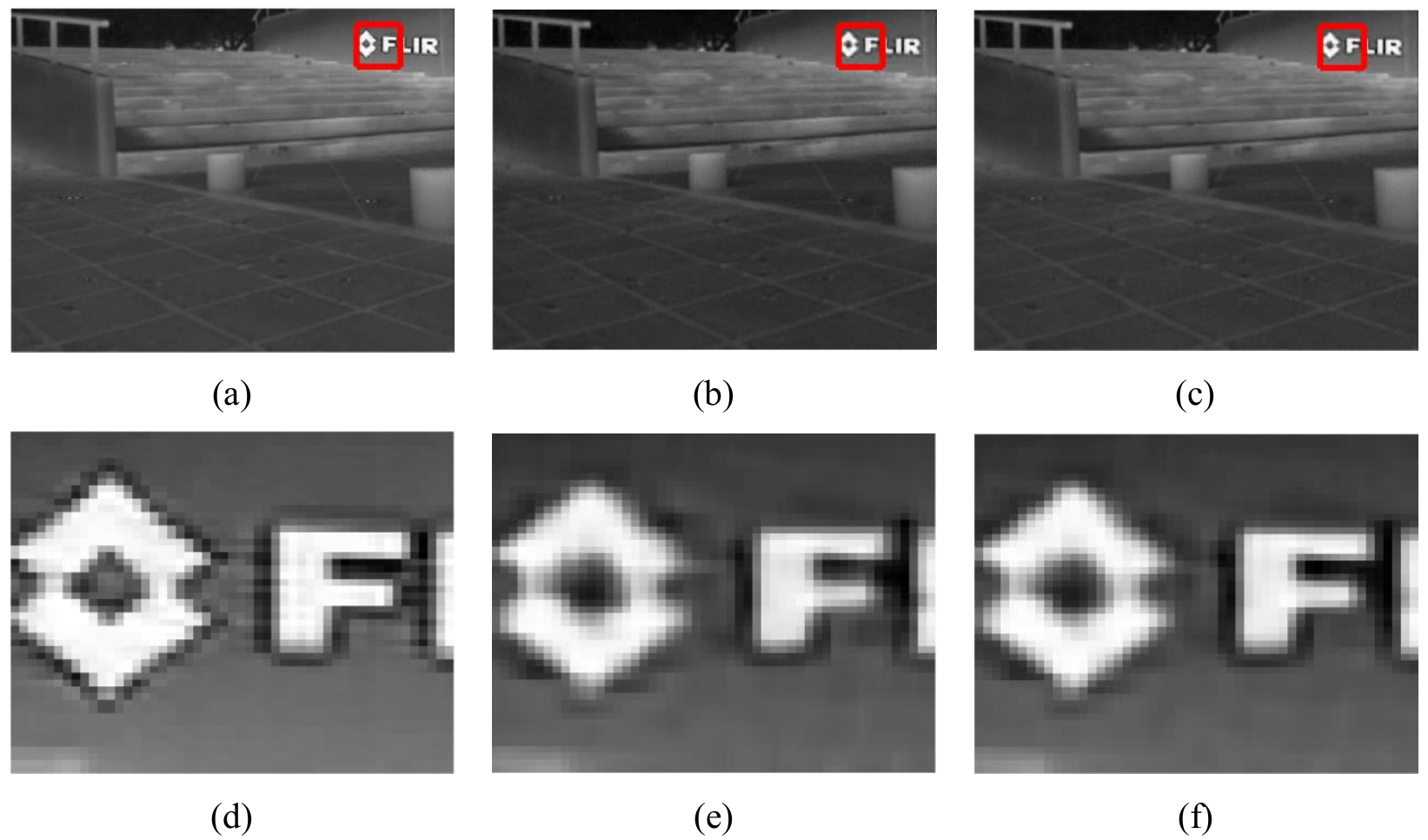}
\end{center}
\caption
{ \label{figure16}
 Results for the Stairs image in Fig.\ref{figure15}: (a) original image, (b) results for OGSATVL1, and (c) results for  OGSATVLp. Enlarged view of the area enclosed by the red squares: (d) original image, (e) results for OGSATVL1, and (f) results for  OGSATVLp.}
\end{figure}

\subsection{Comparison with other TV-based algorithms}
In this section, we compare the method proposed in this study with three other common methods: ITV, ATV, and L0TVPADMM. In the experiment, the Gaussian blur kernel is a $7 \times 7$ kernel with a standard deviation of 5, the salt-and-pepper noise levels are 30\% to 60\%, and the PSNR, SSIM and RE values of the six test images are analyzed. To ensure the maximum PSNR value of each test image, the parameters of the three comparison methods were set separately for each image. The parameter settings of this method were fixed as described above.

The test results are shown in Table \ref{table4}. They show that ITV and ATV have poor robustness against salt-and-pepper noise. Although the overall processing result of L0TVPADMM method is not as good as the proposed method, its PSNR values do not change much when the salt-and-pepper noise level changes from 30\% to 60\%. The PSNR values of the six images are reduced by 1 dB.

The method proposed in this paper has significantly better PSNR, SSIM, and RE values than the other three methods for all six test images under different salt-and-pepper noise levels. However, compared with the L0TVPADMM method, as the salt-and-pepper noise level increases, the PSNR value decreases significantly and the range of change is large.

\begin{table}[htbp]
  \centering
  \caption{Comparison of the proposed and existing methods for a $7 \times 7$ Gaussian blur kernel.}
  \scriptsize
    \begin{tabular}{llccccccccc}
    \toprule
    \multirow{2}[4]{*}{Image} & Noise level & \multicolumn{3}{c}{30\%} & \multicolumn{3}{c}{40\%} & \multicolumn{3}{c}{50\%} \\
\cmidrule{2-11}          & methods & PSNR  & SSIM  & ReE   & PSNR  & SSIM  & ReE   & PSNR  & SSIM  & ReE \\
    \midrule
    \multirow{4}[8]{*}{Truck} & ITV   & 30.0291  & 0.8884  & 0.0616  & 29.5562  & 0.8893  & 0.0651  & 29.1880  & 0.8837  & 0.0691  \\
\cmidrule{2-11}          & ATV   & 35.3620  & 0.9228  & 0.0331  & 33.4723  & 0.9087  & 0.0412  & 31.4261  & 0.8976  & 0.0529  \\
\cmidrule{2-11}          & L0TVPADMM & 37.1920  & 0.9033  & 0.0271  & 36.5800  & 0.8964  & 0.0291  & 35.5870  & 0.8682  & 0.0326  \\
\cmidrule{2-11}          & OGSATVLp & \textbf{39.6943 } & \textbf{0.9472 } & \textbf{0.0205 } & \textbf{38.4550 } & \textbf{0.9391 } & \textbf{0.0236 } & \textbf{36.8715 } & \textbf{0.9253 } & \textbf{0.0283 } \\
    \midrule
    \multirow{4}[8]{*}{Parterre} & ITV   & 36.4479  & 0.9357  & 0.0379  & 36.0066  & 0.9318  & 0.0400  & 34.5629  & 0.9233  & 0.0456  \\
\cmidrule{2-11}          & ATV   & 37.1587  & 0.9376  & 0.0352  & 36.6090  & 0.9321  & 0.0385  & 35.2160  & 0.9221  & 0.0432  \\
\cmidrule{2-11}          & L0TVPADMM & 39.2100  & 0.9290  & 0.0274  & 39.0490  & 0.9347  & 0.0287  & 38.7830  & 0.9260  & 0.0293  \\
\cmidrule{2-11}          & OGSATVLp & \textbf{43.0244 } & \textbf{0.9804 } & \textbf{0.0225 } & \textbf{42.5242 } & \textbf{0.9781 } & \textbf{0.0246 } & \textbf{41.2265 } & \textbf{0.9729 } & \textbf{0.0267 } \\
    \midrule
    \multirow{4}[8]{*}{Stairs} & ITV   & 36.1477  & 0.9353  & 0.0480  & 35.5521  & 0.9303  & 0.0517  & 34.4880  & 0.9225  & 0.0562  \\
\cmidrule{2-11}          & ATV   & 37.2454  & 0.9394  & 0.0426  & 36.3834  & 0.9312  & 0.0469  & 35.4070  & 0.9222  & 0.0517  \\
\cmidrule{2-11}          & L0TVPADMM & 38.8690  & 0.9273  & 0.0350  & 38.6270  & 0.9251  & 0.0360  & 38.3940  & 0.9227  & 0.0370  \\
\cmidrule{2-11}          & OGSATVLp & \textbf{43.0574 } & \textbf{0.9826 } & \textbf{0.0279 } & \textbf{42.1333 } & \textbf{0.9801 } & \textbf{0.0304 } & \textbf{40.6847 } & \textbf{0.9709 } & \textbf{0.0347 } \\
    \midrule
    \multirow{4}[8]{*}{Corridor} & ITV   & 35.1721  & 0.9273  & 0.0544  & 34.5450  & 0.9227  & 0.0581  & 33.8964  & 0.9122  & 0.0676  \\
\cmidrule{2-11}          & ATV   & 35.7739  & 0.9299  & 0.0519  & 34.8115  & 0.9202  & 0.0565  & 33.9243  & 0.9078  & 0.0628  \\
\cmidrule{2-11}          & L0TVPADMM & 38.5150  & 0.9404  & 0.0373  & 38.4240  & 0.9395  & 0.0375  & 38.0550  & 0.9343  & 0.0391  \\
\cmidrule{2-11}          & OGSATVLp & \textbf{42.3201 } & \textbf{0.9813 } & \textbf{0.0304 } & \textbf{41.2782 } & \textbf{0.9775 } & \textbf{0.0331 } & \textbf{40.2169 } & \textbf{0.9719 } & \textbf{0.0368 } \\
    \midrule
    \multirow{4}[8]{*}{Road} & ITV   & 34.1196  & 0.9351  & 0.0443  & 33.5094  & 0.9300  & 0.0472  & 32.6307  & 0.9202  & 0.0520  \\
\cmidrule{2-11}          & ATV   & 34.7909  & 0.9377  & 0.0411  & 33.8026  & 0.9289  & 0.0456  & 32.7069  & 0.9155  & 0.0515  \\
\cmidrule{2-11}          & L0TVPADMM & 38.5640  & 0.9462  & 0.0262  & 38.0270  & 0.9378  & 0.0276  & 37.7320  & 0.9401  & 0.0289  \\
\cmidrule{2-11}          & OGSATVLp & \textbf{42.2739 } & \textbf{0.9845 } & \textbf{0.0217 } & \textbf{41.4298 } & \textbf{0.9825 } & \textbf{0.0236 } & \textbf{40.1804 } & \textbf{0.9781 } & \textbf{0.0268 } \\
    \midrule
    \multirow{4}[8]{*}{Buliding} & ITV   & 30.1056  & 0.8555  & 0.0613  & 29.7495  & 0.8454  & 0.0645  & 29.2963  & 0.8365  & 0.0676  \\
\cmidrule{2-11}          & ATV   & 34.8630  & 0.9071  & 0.0374  & 33.1033  & 0.8860  & 0.0443  & 30.4774  & 0.8579  & 0.0585  \\
\cmidrule{2-11}          & L0TVPADMM & 37.0520  & 0.9258  & 0.0276  & 36.1780  & 0.9121  & 0.0305  & 35.5460  & 0.8971  & 0.0328  \\
\cmidrule{2-11}          & OGSATVLp & \textbf{39.3587 } & \textbf{0.9547 } & \textbf{0.0213 } & \textbf{38.7468 } & \textbf{0.9498 } & \textbf{0.0229 } & \textbf{37.2539 } & \textbf{0.9398 } & \textbf{0.0271 } \\
    \bottomrule
    \end{tabular}%
  \label{table4}%
\end{table}%

Next, we use the three images of Truck, Corridor, and Road to show a visual comparison of the results of the four methods. The images are corrupted with a $ 7 \times 7$ Gaussian blur kernel with a standard deviation of 5 and superimposed with 30\%, 40\% and 50\% of salt-and-pepper noise.

\begin{figure}[htbp!]
\begin{center}
\includegraphics{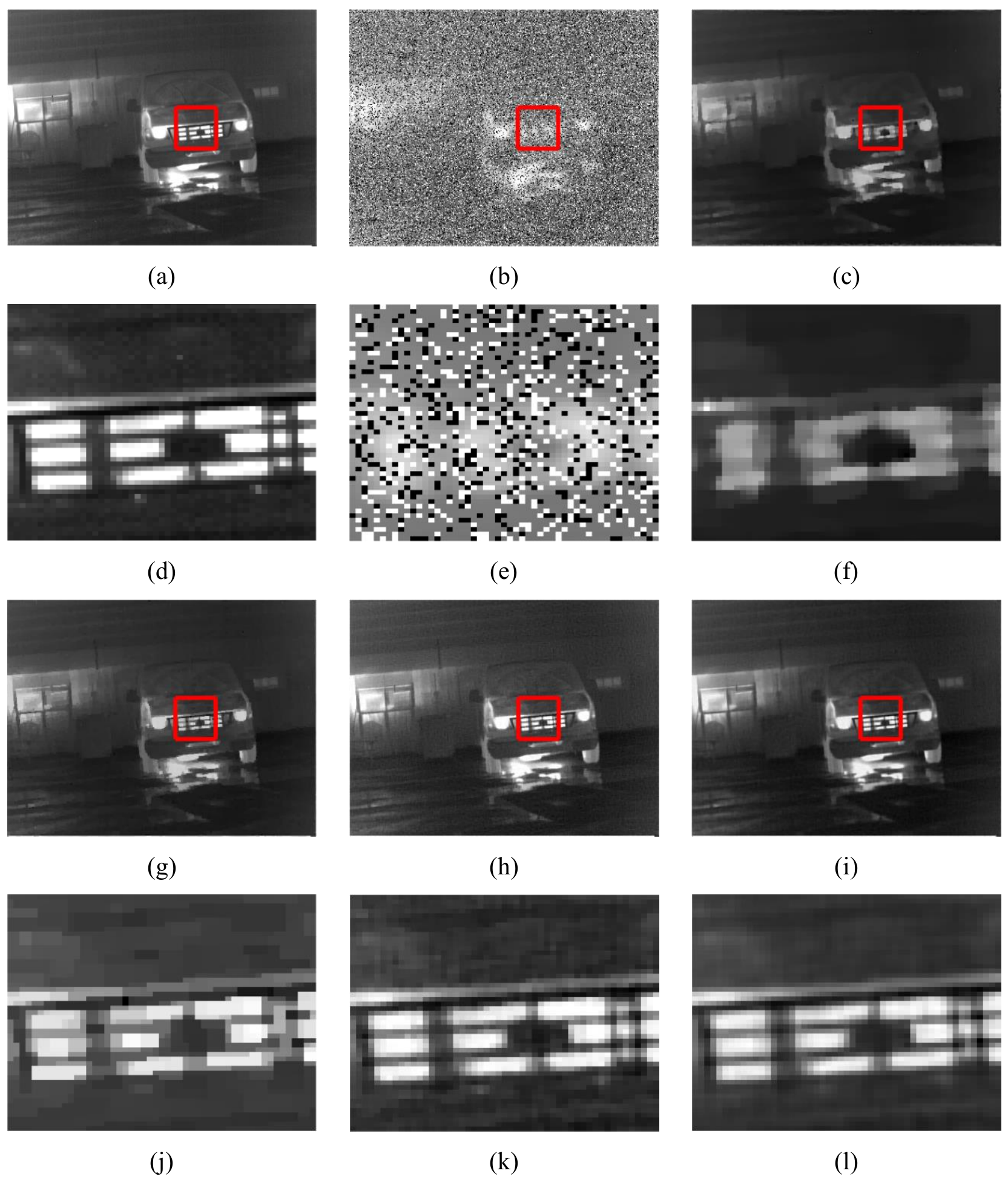}
\end{center}
\caption
{ \label{figure17}
  Comparison of the results obtained by the proposed method and comparison methods on the Truck image: (a) original image, (b) blurred image ($7 \times 7$ Gaussian blur kernel with a standard deviation of 5) with 30\% salt-and-pepper noise, (c) results of ITV, (d)-(f) enlarged regions of the red squares in (a)-(c), respectively;  results of (g) ATV, (h) L0TVPADMM, and (i) proposed method; and (j)-(l) enlarged regions of the red squares in (g)-(i), respectively.}
\end{figure}

The results for the Truck image in Fig.\ref{figure17} show that, in the image after the ITV processing, we can hardly see the horizontal line of the front face of the car in the enlarged image and the whole image is blurred. These results are the worst compared with those of the other three methods. In the results of the ATV method, a part of the horizontal line can be seen in the enlarged view, but the edge of the line is severely stepped, and the straight line is almost always a broken line. In the L0TVPADMM processing result, the line edge step phenomenon is still obvious, and in the smooth image areas (as in the black area in Fig.\ref{figure17}(f)), the block artifact is obvious. Compared with the previous three methods, the processing results of this method have a better recovered line, no obvious staircase artifacts, and good performance in the smooth image area, and there is no block artifact.

\begin{figure}[htbp!]
\begin{center}
\includegraphics{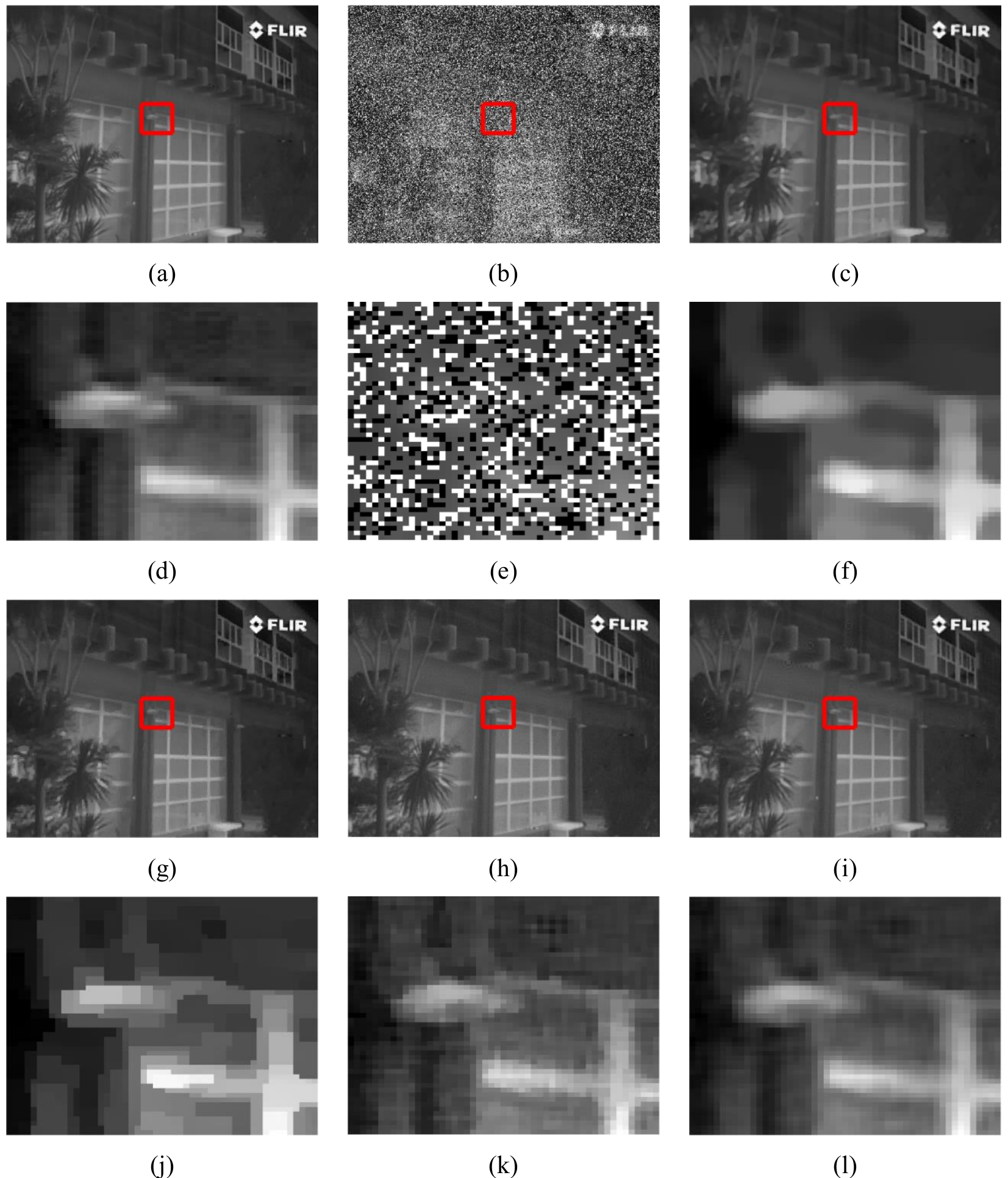}
\end{center}
\caption
{ \label{figure18}
  Comparison of the results obtained by the proposed method and comparison methods on the Corridor image: (a) original image, (b) blurred image ($7 \times 7$ Gaussian blur kernel with a standard deviation of 5) with 40\% salt-and-pepper noise, (c) results of ITV, (d)-(f) enlarged regions of the red squares in (a)-(c), respectively;  results of (g) ATV, (h) L0TVPADMM, and (i) proposed method; and (j)-(l) enlarged regions of the red squares in (g)-(i), respectively.}
\end{figure}

The results for the Corridor image in Fig.\ref{figure18} show that although the enlarged ITV processing results has clear edges, the overall image is still blurred, and the image detail recovery is poor. The results of ATV processing are the worst of the four methods, whether at the edge of the image or in the details. The L0TVPADMM method result has a significant staircase artifact at the edge of the image, and there is a significant block artifact in the magnified view, but the overall image is clearer than those of the ITV and ATV methods. The magnified view shows that the method proposed this paper has good recovery results for the edge of the line and the details of the image. Overall, the recovery results are still the best of the four methods.

\begin{figure}[htbp!]
\begin{center}
\includegraphics{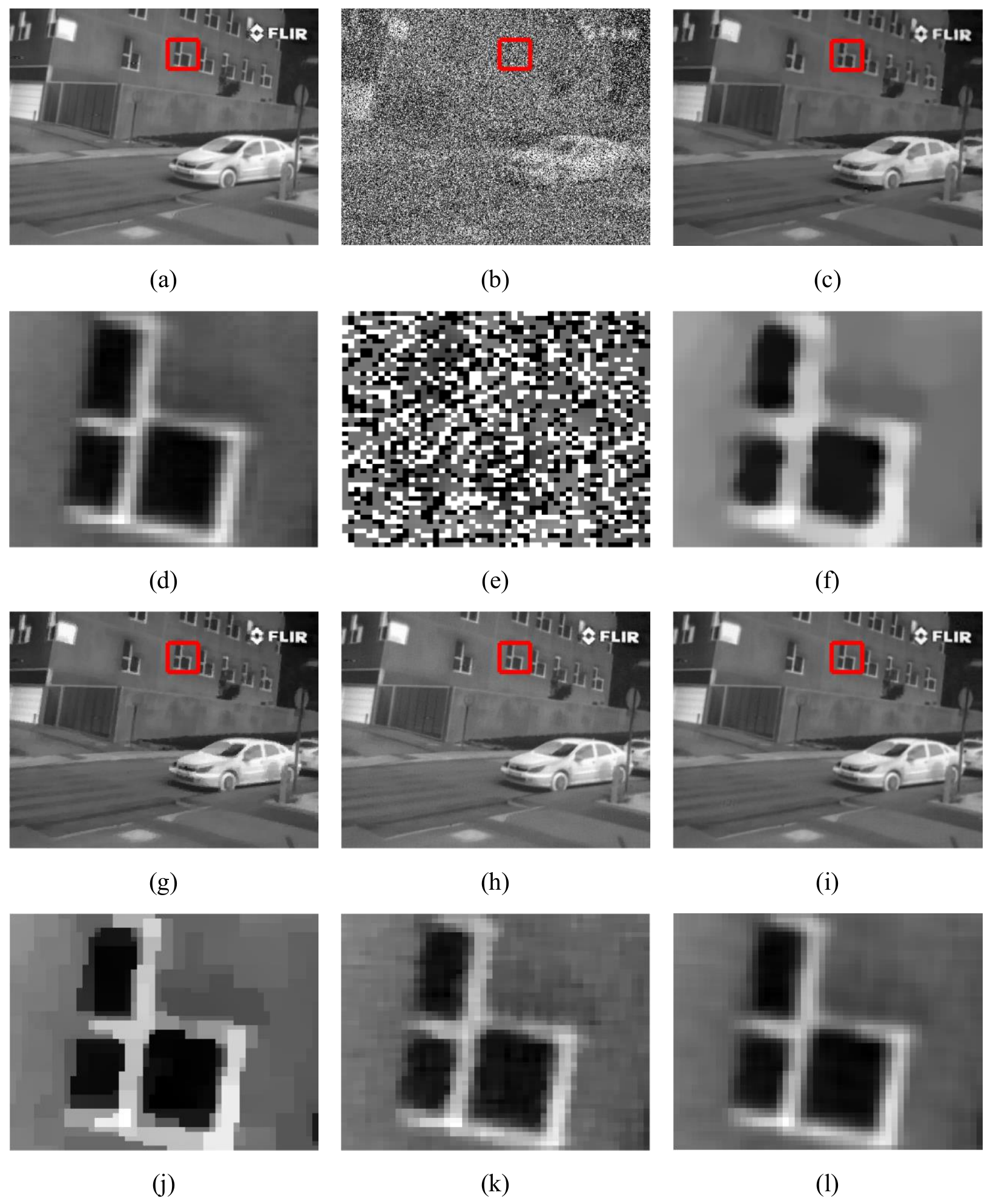}
\end{center}
\caption
{ \label{figure19}
 Comparison of the results obtained by the proposed method and comparison methods on the Road image: (a) original image, (b) blurred image ($7 \times7$ Gaussian blur kernel with a standard deviation of 5) with 50\% salt-and-pepper noise, (c) results of ITV, (d)-(f) enlarged regions of the red squares in (a)-(c), respectively;  results of (g) ATV, (h) L0TVPADMM, and (i) proposed method; and (j)-(l) enlarged regions of the red squares in (g)-(i), respectively. }
\end{figure}

In the results for the Road image in Fig.\ref{figure19}, the enlarged results of the four methods show a large difference in the oblique line processing. Among them, for those of ITV and ATV, although there are clear line edges, there is greater distortion than the original image. Although the L0TVPADMM method restores the original image edge, its staircase artifact is obvious, and the block artifact in the smooth image area is also prominent. With respect to the edges in the image, the smooth regions, and the overall results, the proposed method still performs best of the four methods.


\section{Discussion}
The method proposed in this paper adds the constraint of an Lp quasinorm on the basis of sparse overlapping groups. This because the overlapping group sparse regularization constraints can make full use of the combined neighborhood gradient to improve the differentiation between the smooth regions and edge regions.  Moreover, the Lp quasinorm can improve the characterization of the image gradients. Therefore, combining them obtains a better image reconstruction, both with respect to numerical and visual results.

The OGSATVL1 algorithm uses OGSTV as a regularization term and the L1 norm as the fidelity term. In the experiment, whether the blur kernel is a Gaussian blur kernel or a mean blur kernel, and for different levels of salt-and-pepper noise, the processing results of OGSATVL1 are poor compared with the processing results of the proposed OGSATVLp method. Especially visually, it can be seen that the OGSATVL1 algorithm has a partial staircase artifact in the reconstructed image edges.

The results of the ITV and ATV algorithms on images with a Gaussian blur kernel and salt-and-pepper noise are also poor. The experimental results generally have sharp edges, but the image is blurred or too smooth, resulting in serious image distortion.

In the L0TVPADMM algorithm, the fidelity term is the L0 norm. The numerical results show that its performance for images with a superimposed Gaussian blur kernel and salt-and-pepper noise is lower than the OGSATVL1 method and the proposed OGSATVLp method. The visual assessment shows that the staircase artifacts of the edges and block artifacts of the smooth regions are relatively obvious.

In addition, the fast ADMM was also introduced into the algorithm. This introduction substantially reduces the number of iterations in the process, which improves the operating efficiency of the algorithm.

\section{Conclusion}
This paper proposed a new regularization model for deblurring infrared images containing salt-and-pepper noise, with OGSTV as the regularization term and the Lp quasinorm as the fidelity term. Based on the basic framework of the ADMM and MM methods in the optimization algorithm, steps to accelerate the restart are introduced, which further improve the efficiency of the algorithm. In addition, in the deblurring process, we also regard the difference operator as a convolution operator, so that convolution is used to process the model in the frequency domain, thus avoiding large-scale matrix calculation.

This paper compared the key performance metrics of the proposed method and the OGSATVL1 algorithm under several conditions. The results show that both numerical and visually, the proposed method has obvious advantages. Compared with the ITV, ATV, and L0TVPADMM methods on images blurred with a Gaussian blur kernel and superimposed salt-and-pepper noise, the proposed method also demonstrates superior performance numerically and visually.

The Lp quasinorm-based group sparse method employed in this study can be easily extended to other regularization models, such as an Lp quasinorm-based generalized total grouping sparse method; we will continue to perform these extensions in our follow-up work.

% \disclosures
\subsection*{Disclosures}
The authors declare no conflict of interest.

\acknowledgments
This research was funded by the National Natural Science Foundation of China (Nos. 61571096, 61775030, 61575038), the Scientific and Technological Research Program of Chongqing Municipal Education Commission (No. KJ1729409), the Natural Science Foundation of Fujian Province (No. 2015J01270), the Education and Scientific Research Foundation of Education Department of Fujian Province for Middle-aged and Young Teachers (Nos. JAT170352, JT180309, JT180310, JT180311), the Foundation of Fujian Province Great Teaching Reform (No. FBJG20180015), the Foundation of Department of Education of Guangdong Province (No. 2017KCXTD015), the Chongqing Educational Science Planning Subject (No. 2015-ZJ-009), and the Open Foundation of Digital Signal and Image Processing Key Laboratory of Guangdong Province (No. 2017GDDSIPL-01).

%%%%% References %%%%%

\bibliography{report}   % bibliography data in report.bib
\bibliographystyle{spiejour}   % makes bibtex use spiejour.bst

%%%%% Biographies of authors %%%%%

%\vspace{2ex}\noindent\textbf{First Author} is an assistant professor at the University of Optical Engineering. He received his BS and MS degrees in physics from the University of Optics in 1985 and 1987, respectively, and his PhD degree in optics from the Institute of Technology in 1991.  He is the author of more than 50 journal papers and has written three book chapters. His current research interests include optical interconnects, holography, and optoelectronic systems. He is a member of SPIE.

%\vspace{1ex}
%\noindent Biographies and photographs of the other authors are not available.

\listoffigures
\listoftables

\end{spacing}
\end{document}